\documentclass{article}
\usepackage{iclr2027_conference,times}

\usepackage{amsmath,amsfonts,bm}

\def\eqref#1{equation~\ref{#1}}

\def\1{\bm{1}}

\DeclareMathAlphabet{\mathsfit}{\encodingdefault}{\sfdefault}{m}{sl}
\SetMathAlphabet{\mathsfit}{bold}{\encodingdefault}{\sfdefault}{bx}{n}

\usepackage{amsmath,amssymb}
\usepackage{booktabs}
\usepackage{graphicx}
\usepackage{subcaption}
\usepackage{wrapfig}
\usepackage{array}
\usepackage{tabularx}
\usepackage{multirow}
\usepackage[table]{xcolor}
\usepackage{algorithm}
\usepackage{algpseudocode}
\floatstyle{ruled}
\restylefloat{algorithm}
\usepackage{hyperref}
\usepackage{url}
\newcolumntype{Y}{>{\raggedright\arraybackslash}X}

\title{VAttnPO: Visual Attention-Guided Policy Optimization}
\author{
\begin{tabular}[t]{@{}p{0.48\textwidth}@{}p{0.48\textwidth}@{}}
\textbf{Lihao Zheng} & \textbf{Zhenwei Shao} \\
\mdseries School of Computer Science and Technology, & \mdseries School of Computer Science and Technology,  \\
\mdseries Hangzhou Dianzi University & \mdseries Hangzhou Dianzi University \\[1.2ex]
\textbf{Zhou Yu} & \textbf{Yan Yang} \\
\mdseries School of Computer Science and Technology,  & \mdseries School of Computer Science and Technology,  \\
\mdseries Hangzhou Dianzi University & \mdseries Hangzhou Dianzi University \\[1.2ex]
\textbf{Haoyi Sun} & \textbf{Qian Wang} \\
\mdseries Base Model, Li Auto & \mdseries Base Model, Li Auto \\[1.2ex]
\textbf{Ning Mao} & \textbf{Pan Zhou} \\
\mdseries Base Model, Li Auto & \mdseries Base Model, Li Auto \\
\end{tabular}
}

\newcommand{\best}[1]{\textbf{#1}}
\iclrfinalcopy
\begin{document}
\maketitle
\lhead{Under review as a conference paper at ICLR 2027}

\begin{abstract}
Multimodal reinforcement learning with verifiable rewards commonly assigns one sequence advantage to an entire response. This outcome-level signal distinguishes overall response quality but leaves changes in visual-evidence use across generated tokens implicit. We introduce \textbf{VAttnPO} (Visual Attention-Guided Policy Optimization), which uses answer-conditioned visual attention to refine token-level credit while preserving the verifier-derived trajectory-mean advantage. A frozen old policy constructs a visual reference by teacher forcing the known correct answer and produces a visual distribution for each rollout token. VAttnPO measures their agreement in visual-attention space using Jensen--Shannon divergence, rather than reading guidance from teacher output-token distributions or introducing an attention-matching objective. Trajectory centering, relative-position weighting, and credit balancing turn this agreement into a zero-mean, position-aware correction to the GRPO sequence advantage. Token intervention, visual-index permutation, and trajectory analysis provide diagnostic evidence linking this signal to model behavior, visual correspondence, and generation position. On Qwen3-VL-8B-Instruct, VAttnPO raises six-task image AVE from 70.97 with GRPO to 73.50 and scores higher on every reported structured, video, and MME metric. Gains over GRPO also hold on Qwen2.5-VL-7B-Instruct and InternVL3-8B-Instruct.
\end{abstract}

\section{Introduction}

Multimodal large language models are advancing rapidly, spanning tool-integrated reasoning, streaming perception, multi-image grounding, policy optimization, and large-scale evaluation~\citep{chen2026mindwatchersmartermultimodaltoolintegrated,shen2026evolvingtoolusercreator,chen2026streamingclawtechnicalreport,11464665,han2025copoconsistencyawarepolicyoptimization,Xu_2025_ICCV}. Reinforcement learning with verifiable rewards (RLVR) uses automatic feedback, such as answer correctness and format compliance, to improve reasoning. Methods such as GRPO estimate policy advantages from relative rewards among multiple responses to the same prompt, avoid training a separate value network, and have expanded from text reasoning to vision-language models~\citep{shao2024deepseekmath,guo2025deepseek,huang2025visionr1incentivizingreasoningcapability,shen2025vlm,meng2025mm,chen2025r1-v}. Yet the environment usually returns only one scalar reward per response, and the resulting sequence advantage is applied to the entire generated trajectory. Every token in a response therefore receives the same advantage coefficient, leaving differences in credit within the trajectory implicit, even when individual tokens rely on different visual evidence.

This granularity gap is especially important in multimodal reasoning: a correct answer does not imply consistent use of relevant visual regions, and equally rewarded responses may follow different visual reasoning paths. Existing fine-grained approaches rely on preference or process supervision~\citep{yu2024rlhf,wang2025visualprm}, or read dense guidance from teacher- and privileged-context output-token distributions~\citep{agarwal2024onpolicy,yang2026selfdistilledrlvr,pan2026rlcsdreinforcementlearningcontrastive}. A vision-language model also produces attention over visual inputs at every generated token, providing an internal readout distinct from output behavior. We treat visual attention as a proxy to be evaluated through behavioral diagnostics and controlled training comparisons, rather than as a causal attribution or certificate of correctness~\citep{liu2026seeing,zhao2026tell,lu2026reallocating,yang2026mitigating}. This leads to our central question: \emph{Can token-varying internal visual statistics refine credit within a trajectory while preserving the verifier-derived trajectory-mean advantage?}

Attention magnitude itself has no intrinsic task direction, so directly rewarding larger attention does not address this question. We instead use the visual distribution formed by the frozen old policy $\pi_{\theta_{\mathrm{old}}}$ under the known correct answer as a reference and measure each generated token's agreement with it. Token intervention tests whether attention rank tracks behavioral sensitivity; visual-index permutation tests whether correct--incorrect separation depends on patch identity rather than the reference's marginal shape; and trajectory analysis tests whether this separation varies with generation position. These diagnostics provide evidence for a behaviorally relevant token proxy, reference-patch correspondence, and position-dependent variation. The position-dependent alignment profile further motivates us to adopt a smooth, bounded relative-position sigmoid and evaluate this design through controlled ablations.

Motivated by this evidence, we introduce \textbf{VAttnPO}. Within each update, the frozen old policy both samples responses and produces the visual reference by teacher forcing the correct answer, placing reference and rollout attention in the same parameter state; the answer supplies conditioning context without process labels or region annotations. VAttnPO converts token-wise visual similarity into trajectory-centered credit, modulates it with a smooth bounded function of relative position, and balances it into a zero-sum correction. Adding this correction to the GRPO sequence advantage allows token coefficients to vary while preserving the verifier-derived trajectory-mean advantage.

To summarize, our contributions are as follows:
\begin{itemize}
\item \textbf{Diagnostic Evidence.} We identify behavioral, spatial, and positional structure in internal visual attention that motivates token-level credit differentiation.
\item \textbf{Visual Attention-Guided Credit.} VAttnPO derives a token-level correction from answer-conditioned visual attention through similarity centering, relative-position weighting, and sign-preserving balancing, varying credit within a response while preserving the trajectory-mean advantage.
\item \textbf{Empirical Results.} On Qwen3-VL-8B-Instruct, VAttnPO improves AVE over six image tasks by 2.53 points over GRPO and records higher scores on every reported metric for structured understanding, video understanding, and MME.
\end{itemize}

\section{Related Work}
\subsection{Multimodal RLVR and Credit Assignment}
GRPO and DeepSeek-R1 establish group-relative RLVR~\citep{shao2024deepseekmath,guo2025deepseek}; Vision-R1, VLM-R1, MM-Eureka, R1-V, and Video-R1 extend it to image and video reasoning~\citep{huang2025visionr1incentivizingreasoningcapability,shen2025vlm,meng2025mm,chen2025r1-v,Video-R1}, while SAPO further modifies policy optimization~\citep{gao2025softadaptivepolicyoptimization}. These methods retain response-level verifiable outcomes as task supervision, and standard GRPO shares one sequence advantage across a completion. Preference learning and process supervision offer finer feedback but generally require preference data, step labels, a process verifier, or an additional reward model~\citep{yu2024rlhf,sun2024aligning,li2023silkie,lightman2024let,wang2024math,wang2025visualprm}. VAttnPO adds none of these sources and leaves verifier rewards unchanged; it only redistributes the resulting sequence advantage within each response.

\subsection{On-policy Distillation and Self-Distilled RL}
GKD and OPD provide token-level targets from teacher output distributions on student trajectories~\citep{agarwal2024onpolicy,lu2025onpolicydistillation}, while OPSD uses the same model as teacher and student under different contexts~\citep{zhao2026selfdistilledreasoneronpolicyselfdistillation}. RLSD reweights verifier-derived advantages using token-likelihood relations between original and correct-answer contexts~\citep{yang2026selfdistilledrlvr}, whereas RLCSD contrasts policy distributions under correct and incorrect hints~\citep{pan2026rlcsdreinforcementlearningcontrastive}. All read reference behavior from output-token distributions. VAttnPO instead reads an answer-conditioned visual-attention distribution and constructs a zero-sum correction to verifier-derived advantages, without an output-distribution or attention-matching objective; verifiable rewards remain the source of task supervision.

\subsection{Attention-Derived Supervision and Visual Credit}
Attention transfer and distillation match teacher attention maps for knowledge transfer~\citep{zagoruyko2016paying,zhou2025attention}, whereas diagnostic and intervention work studies links between visual attention, grounding, hallucination, and behavior~\citep{liu2026seeing,zhao2026tell,lu2026reallocating,yang2026mitigating}. These studies show that attention is neither a causal explanation nor a certificate of correctness, yet it is not behaviorally arbitrary. VAttnPO neither increases attention at prescribed locations nor matches attention maps: the same frozen old policy produces the answer-conditioned reference and rollout attention, and the derived statistics are used only to construct the token-level GRPO correction.

\section{Pilot Study}
\label{sec:pilot}
We examine whether visual attention tracks behavioral sensitivity, whether correctness-related separation depends on visual-patch correspondence, and how this separation varies over completion progress. The answer-conditioned reference $P_{\mathrm{gt}}$ pools the frozen model's attention over correct-answer queries and normalizes it over visual patches (Equation~(2)); lower JSD indicates that a rollout token's visual distribution is closer to this reference.
We conduct three targeted pilot studies to examine these questions. All studies use layer 20 of Qwen3-VL-8B-Instruct, the same fixed 200 examples, and $G=8$ responses per input. The analysis includes 1,429 valid trajectories, comprising 771 correct and 658 incorrect responses. Correct--incorrect statistics are aggregated within input before inputs are weighted equally, with input-cluster bootstrap intervals; Appendix~\ref{app:diagnostics} provides additional protocol details.

\begin{figure}[t]
\centering
\begin{minipage}[t]{0.32\linewidth}\centering
\includegraphics[width=\linewidth]{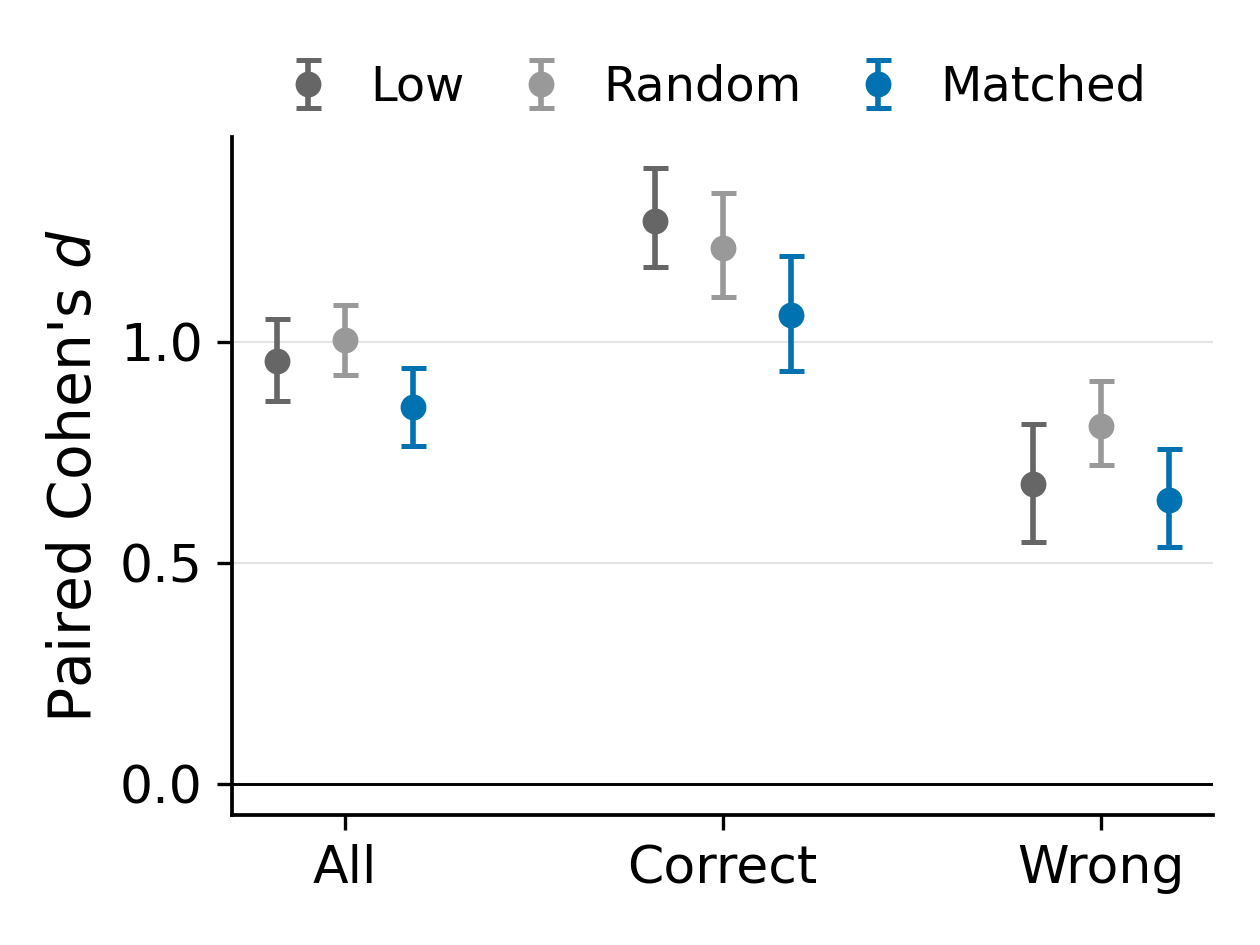}\\[-1mm]
\small (a) Token intervention
\end{minipage}\hfill
\begin{minipage}[t]{0.32\linewidth}\centering
\includegraphics[width=\linewidth]{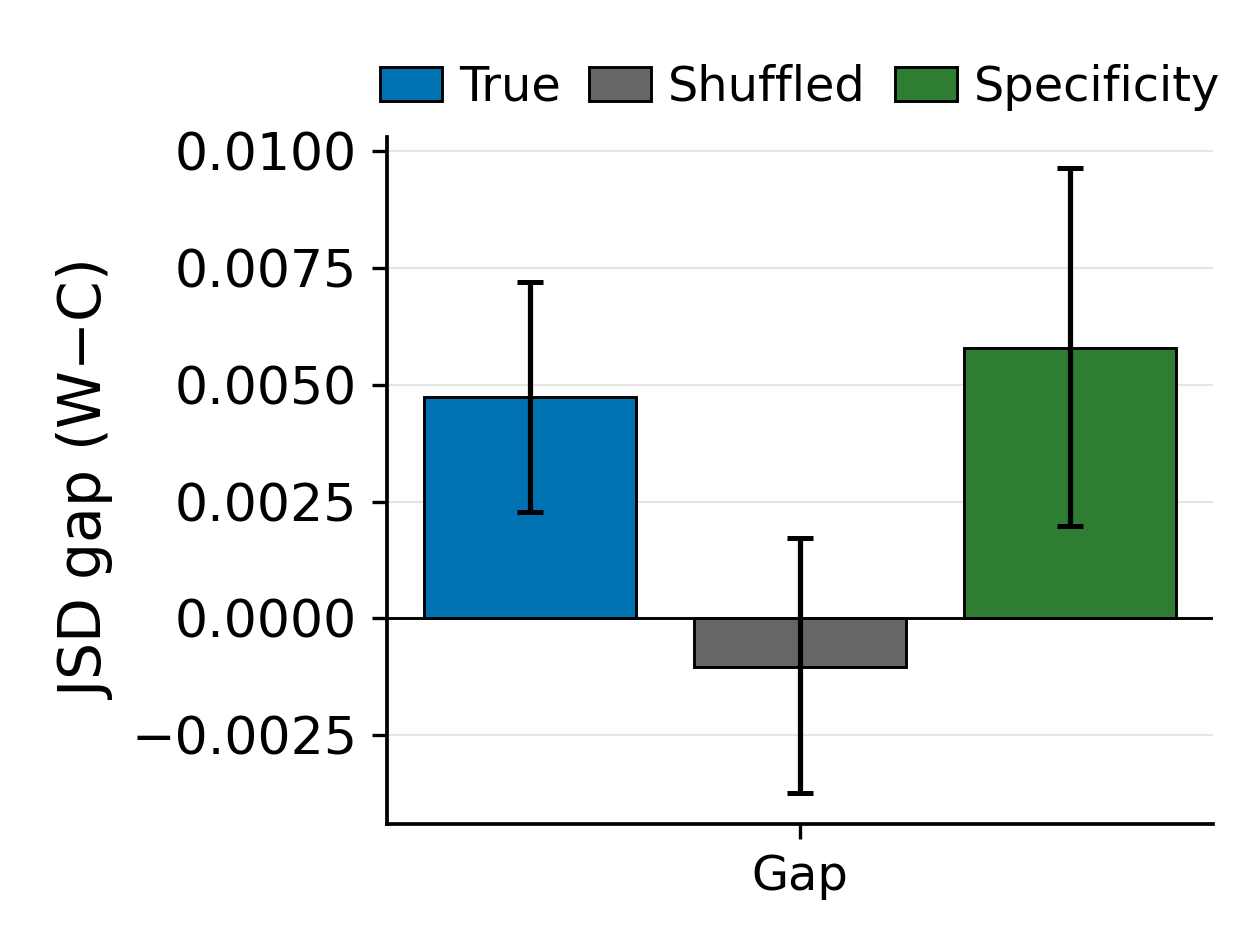}\\[-1mm]
\small (b) GT vs. shuffled reference
\end{minipage}\hfill
\begin{minipage}[t]{0.32\linewidth}\centering
\includegraphics[width=\linewidth]{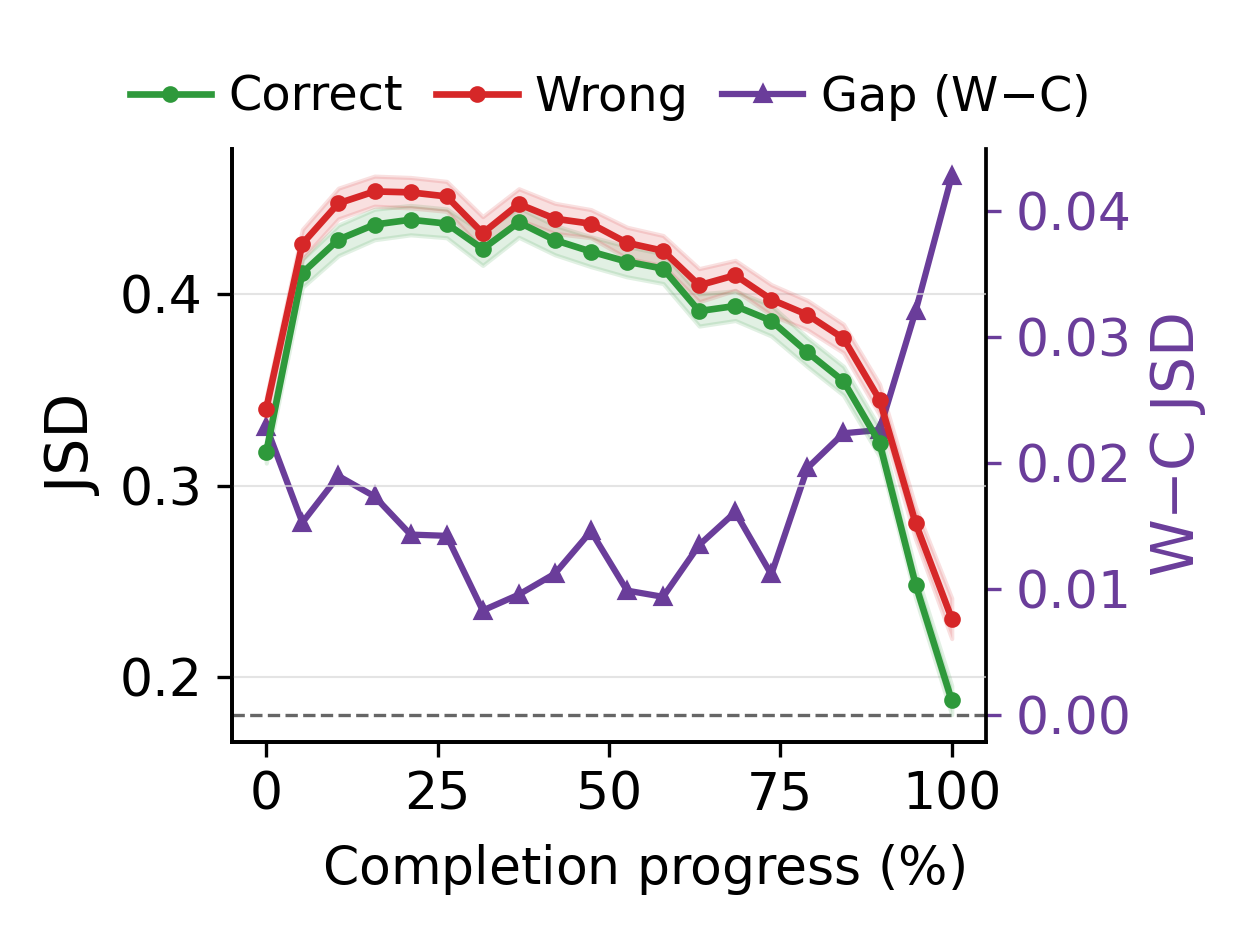}\\[-1mm]
\small (c) Alignment profiles
\end{minipage}
\caption{Diagnostic evidence from visual attention. (a) Paired Cohen's $d$ for High tokens against Low, Random, and Position-matched controls. (b) Wrong-minus-correct JSD gaps under the correct-answer and patch-shuffled references. (c) Mean JSD for correct and incorrect responses and their wrong-minus-correct gap over completion progress. 
The final point denotes the last valid marker. Error bars in (a) and (b), and shaded bands around the correct and incorrect curves in (c), show confidence intervals obtained by bootstrap resampling at the example level.
}
\label{fig:motivation}
\end{figure}

\paragraph{Token Intervention.}
We first test whether attention rank tracks behavioral sensitivity. Using the top 20\% of visual patches under $P_{\mathrm{gt}}$ as the reference region, we rank tokens by the preceding query's normalized attention to that region and form equal-size high, low, random, and position-matched sets. We then mask the selected tokens' complete self-attention key--value columns. Over all responses, high-attention tokens cause larger drops in the original answer log-probability than all controls, with paired Cohen's $d=0.85$--$1.01$ and intervals excluding zero; high and position-matched sets have similar mean positions. Figure~\ref{fig:motivation}(a) additionally reports the correct and incorrect response subsets, and its legend abbreviates Position-matched as Matched. Attention rank therefore distinguishes tokens with different intervention effects even after matching average position.

\paragraph{GT Versus Shuffled Reference.}
We compare rollout-token JSD to the correct-answer reference and to a visual-index-permuted version that preserves its attention values and entropy but breaks visual correspondence. Incorrect trajectories are farther from the correct-answer reference on average, whereas the shuffled-reference gap is near zero; the interval for the difference between these gaps excludes zero. In Figure~\ref{fig:motivation}(b), True denotes the wrong-minus-correct gap under the correct-answer reference, Shuffled denotes the gap under the permuted reference, and Specificity is the True gap minus the Shuffled gap. Correct--incorrect separation therefore depends on reference-patch correspondence rather than only the marginal distribution of reference values.

\paragraph{Alignment Over Completion Progress.}
Correct responses have lower mean JSD than incorrect responses at every displayed position. The wrong-minus-correct gap varies non-monotonically and is largest at the final valid token, showing position-dependent alignment without implying a general late-reasoning mechanism (Figure~\ref{fig:motivation}(c)).

These diagnostics motivate using answer-conditioned visual attention to differentiate tokens and examining its variation over generation progress. The position-dependent alignment profile motivates us to adopt a smooth, bounded relative-position sigmoid and evaluate this design through controlled ablations in Section~\ref{sec:ablation}. Centering and credit balancing preserve the trajectory mean of the verifier-derived advantage.

\section{Method}
VAttnPO adds a token-level visual correction to the GRPO sequence advantage while preserving its trajectory mean. The verifiable reward and its group-relative advantage evaluate the complete response, while the internal visual signal redistributes that advantage within the response.

Constructing this correction requires explicit constraints: the answer-conditioned reference supplies task direction, trajectory centering removes response-wide offset, bounded position weights modulate local credit, and credit balancing restores zero mean while retaining nonzero credit signs. The three stages construct reference and rollout attention, compute centered credit, and form position-aware token advantages.

Concretely, $P_{\mathrm{gt}}$ and $Q_{i,t}$ produce visual similarity $s_{i,t}$; trajectory centering yields $r_{i,t}$; position modulation forms $u_{i,t}$; and credit balancing produces $d_{i,t}$. The final coefficient $\widetilde A_{i,t}=\hat A_i+\alpha d_{i,t}$ retains $\hat A_i$ as the verifier-derived response-level evaluation while $d_{i,t}$ controls its local allocation within the response.

\begin{figure}[t]
\centering
\includegraphics[width=0.98\linewidth]{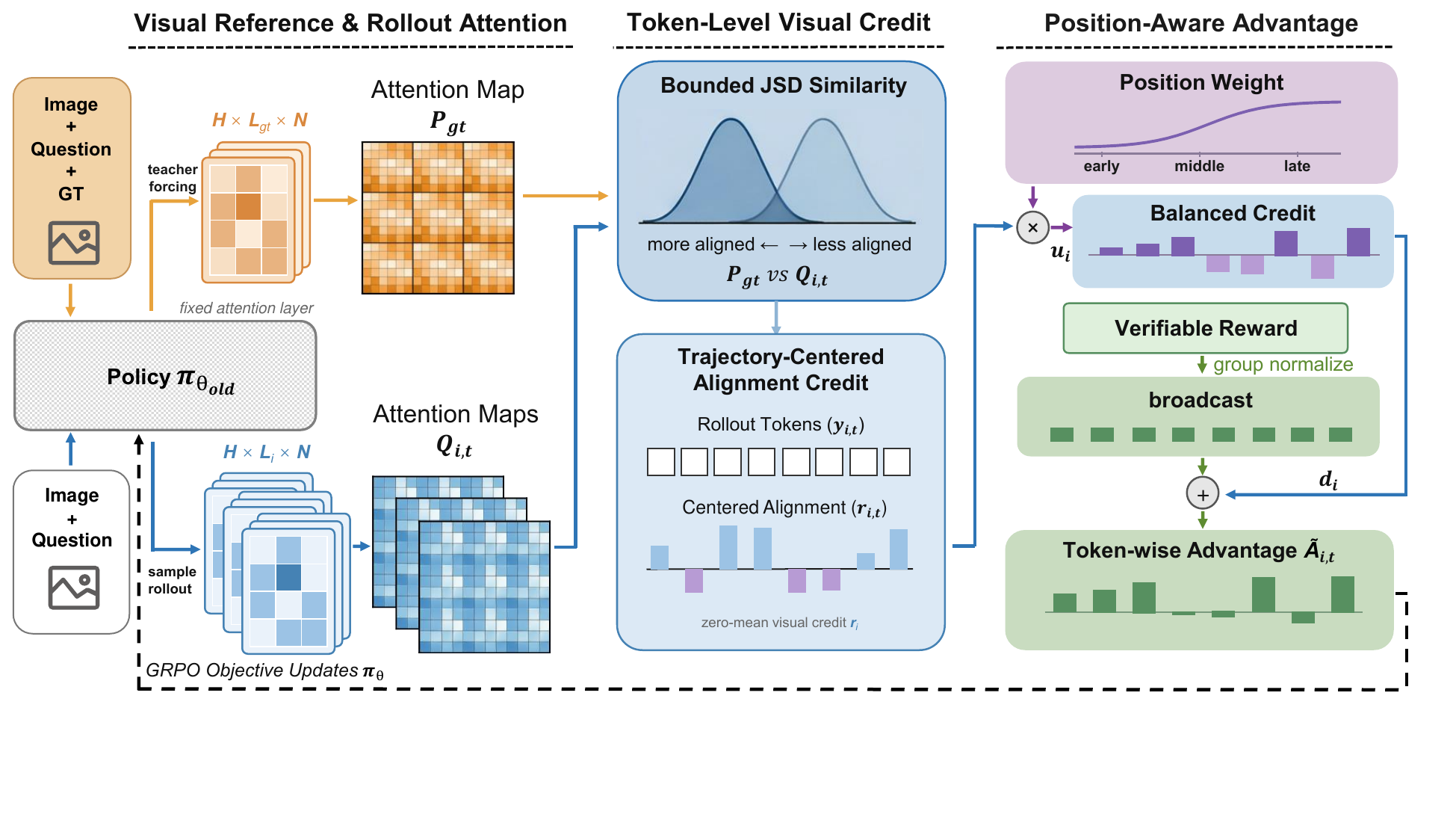}
\caption{VAttnPO overview. Within each update, the frozen old policy extracts visual distributions by teacher forcing the correct answer and by processing sampled responses. Their similarity is centered within each trajectory, weighted by relative position, and balanced into a token correction of the sequence advantage.}
\label{fig:method}
\end{figure}

\subsection{Preliminaries}
Given multimodal input $x$, the frozen old policy $\pi_{\theta_{\mathrm{old}}}$ samples $G$ responses $y_i=(y_{i,1},\ldots,y_{i,L_i})$, where $L_i$ is the valid completion length. The old policy remains fixed within an update, placing sampling probabilities, response attention, and the visual reference obtained under the correct answer in the same parameter state. The policy $\pi_{\mathrm{ref}}$ used for KL regularization is a separate frozen model with a different role.

The verifiable reward $R_i$ gives the group-relative advantage
\begin{equation}
\hat A_i=
\frac{R_i-\operatorname{mean}(R_{1:G})}
{\operatorname{std}(R_{1:G})+\epsilon}.
\tag{1}
\end{equation}
Standard GRPO broadcasts $\hat A_i$ to every valid token, so token-level differences arise only from likelihood ratios, clipping states, and score functions rather than explicit credit. VAttnPO leaves $R_i$ and the group-normalized sequence advantage $\hat A_i$ unchanged. It replaces the broadcast $\hat A_i$ with token-varying $\widetilde A_{i,t}$ having the same trajectory mean. Figure~\ref{fig:method} summarizes the path from visual attention to these coefficients.

\subsection{Stage 1: Visual Reference and Rollout Attention}
For each input, we run two forward paths through the same $\pi_{\theta_{\mathrm{old}}}$. The reference branch teacher-forces the known correct answer $y^{\mathrm{gt}}$, while the response branch reads attention from responses sampled by that old policy. Both paths share model parameters, input visual tokenization, and the target attention layer. 
Therefore, their comparison focuses on the distribution of visual attention generated under the same strategy state across two text conditions, rather than on representation drift across models or updates.

At the fixed layer, we retain only post-softmax attention from textual queries to the $N$ visual-patch keys. Let $\mathbf a^{\mathrm{gt}}_{m,h,:}\in\mathbb R_+^N$ and $\mathbf a^{(i)}_{t,h,:}\in\mathbb R_+^N$ denote visual attention at correct-answer token $m$ and rollout token $t$ for head $h$. The reference branch pools over $L_{\mathrm{gt}}$ valid answer queries and $H$ heads, whereas the rollout branch retains one distribution for every valid generation position. Normalizing the pooled weights over visual patches gives
\begin{equation}
P_{\mathrm{gt}}=\operatorname{Norm}\!\Big(\tfrac{1}{L_{\mathrm{gt}}H}{\textstyle\sum_{m=1}^{L_{\mathrm{gt}}}\sum_{h=1}^{H}}\mathbf a^{\mathrm{gt}}_{m,h,:}\Big),\qquad
Q_{i,t}=\operatorname{Norm}\!\Big(\tfrac{1}{H}{\textstyle\sum_{h=1}^{H}}\mathbf a^{(i)}_{t,h,:}\Big).
\tag{2}
\end{equation}
All responses to an input share $P_{\mathrm{gt}}$ and are therefore compared with the same visual reference. The reference is obtained from the old policy under the known correct answer; the answer supplies conditioning context but no region supervision. After generation, VAttnPO assigns $Q_{i,t}$ to token $t$ at the same index, enabling relative position normalization by the final valid length without altering sampling. Appendix~\ref{app:implementation} outlines attention aggregation and the query-index distinction from the intervention in Section~\ref{sec:pilot}. Appendix~\ref{app:inference-layer-localization} examines the layer-wise profile of correctness-related visual-attention separation across three backbones, while Appendix~\ref{app:layerwise-examples} illustrates spatial attention patterns across layers.

\subsection{Stage 2: Token-Level Visual Credit}
Both $Q_{i,t}$ and $P_{\mathrm{gt}}$ are probability distributions over the same visual-patch support. We use Jensen--Shannon divergence (JSD) because it is symmetric, always finite, and bounded above by the sample-independent constant $\ln 2$. These properties provide a stable mapping from distance to $[0,1]$ without requiring identical nonzero support. The resulting bounded similarity is
\begin{equation}
s_{i,t}=1-\frac{\operatorname{JSD}(Q_{i,t}\Vert P_{\mathrm{gt}})}{\ln 2}\in[0,1].
\tag{3}
\end{equation}
Absolute similarity can vary with task content, the size of the visual support, and the concentration of the reference distribution. Treating $s_{i,t}$ directly as an added reward would carry the similarity offset shared by a response into local credit and could duplicate sequence-level differences. We therefore subtract the mean over valid generated tokens:
\begin{equation}
r_{i,t}=s_{i,t}-\frac1{L_i}\sum_{u=1}^{L_i}s_{i,u},
\qquad
\sum_{t=1}^{L_i}r_{i,t}=0.
\tag{4}
\end{equation}
The sign of $r_{i,t}$ indicates whether token $t$ is more or less aligned than its trajectory average. Centering removes the response-wide offset and yields zero-sum local credit. It preserves every within-trajectory similarity difference because $r_{i,t}-r_{i,u}=s_{i,t}-s_{i,u}$, but does not rescale by variance, so different responses retain their own credit dispersion.

\subsection{Stage 3: Position-Aware Advantage Construction}
\paragraph{Position Modulation.}
We use a smooth, bounded position weight to modulate visual credit over completion progress without directly scaling $\hat A_i$. For $L_i\geq2$, let $p_{i,t}=(t-1)/(L_i-1)$ denote relative completion position. With $\kappa>0$, location parameter $\mu$, and $0<g_{\min}\leq1$, the endpoint-normalized sigmoid, position weight, and weighted credit are
\begin{equation}
\begin{aligned}
S_\kappa(p)
&=\frac{\sigma(\kappa(p-\mu))-\sigma(-\kappa\mu)}
{\sigma(\kappa(1-\mu))-\sigma(-\kappa\mu)},\\
g^{\mathrm{pos}}_{i,t}
&=g_{\min}+2(1-g_{\min})S_\kappa(p_{i,t}),\\
u_{i,t}
&=g^{\mathrm{pos}}_{i,t}r_{i,t}.
\end{aligned}
\tag{5}
\end{equation}
Here, $\mu$ locates the sigmoid transition, $\kappa$ controls its steepness, and $g_{\min}$ sets the bounded weight range from $g_{\min}$ to $2-g_{\min}$. When $\mu=0.5$, the midpoint weight is one. The construction uses only valid length and relative position, placing different-length responses on the same smooth progress coordinate; for $L_i=1$, we use unit weight. Position alone creates no independent advantage and changes only the magnitude of existing positive and negative visual credits.

\paragraph{Credit Balancing.}
Although $\sum_t r_{i,t}=0$, position modulation generally gives $\sum_tu_{i,t}\neq0$. Directly re-centering $\mathbf u_i$ would restore zero sum but could change a local credit's sign. Let $[z]_+=\max(z,0)$. We compute positive and negative weighted-credit totals separately and scale both sides to the smaller total:
\begin{equation}
P_i^+=\sum_t[u_{i,t}]_+,
\qquad
N_i^-=\sum_t[-u_{i,t}]_+,
\qquad
C_i=\min(P_i^+,N_i^-).
\tag{6}
\end{equation}
\begin{equation}
d_{i,t}=C_i\left(\frac{[u_{i,t}]_+}{P_i^+}-\frac{[-u_{i,t}]_+}{N_i^-}\right),
\qquad \sum_t d_{i,t}=0.
\tag{7}
\end{equation}
$C_i$ is the largest credit mass that both signs can retain, so the smaller side remains unchanged and only the larger side is contracted. In a nondegenerate trajectory, both scaling factors lie in $(0,1]$; hence each $d_{i,t}$ retains the sign of its corresponding $u_{i,t}$ without increasing its magnitude, while the positive and negative totals cancel exactly. If either total does not exceed the numerical threshold $\tau$, the entire $\mathbf d_i$ is set to zero.

\paragraph{Policy Update.}
The final token advantage uses correction strength $\alpha$:
\begin{equation}
\widetilde A_{i,t}=\hat A_i+\alpha d_{i,t},
\qquad
\sum_{t=1}^{L_i}\widetilde A_{i,t}=L_i\hat A_i.
\tag{8}
\end{equation}
Thus, $\alpha d_{i,t}$ changes token coefficients within a response, while $\sum_t d_{i,t}=0$ preserves the trajectory mean of the verifier-derived sequence advantage. All attention, similarity, position-weight, and balance statistics are stop-gradient. During training, $\widetilde A_{i,t}$ directly replaces the broadcast $\hat A_i$ in the clipped GRPO objective, while likelihood ratios, clipping, and KL regularization against the separate frozen policy $\pi_{\mathrm{ref}}$ remain unchanged. Appendix~\ref{app:implementation} gives the full objective, degenerate branches, and Algorithm~\ref{alg:vattnpo}; Appendix~\ref{app:properties} provides the corresponding properties and derivations.

\section{Experiments}
\label{sec:experiments}
\subsection{Experimental Setup}
All primary experiments start from Qwen3-VL-8B-Instruct~\citep{bai2025qwen3} and train directly, without SFT warm-up, on a fixed set of 168,460 multiple-choice image--video examples from Video-R1-260k~\citep{Video-R1}. Except in the training-step ablation, the training budget is 400 optimization steps. We post-train separate models with GRPO~\citep{shao2024deepseekmath}, SAPO~\citep{gao2025softadaptivepolicyoptimization}, On-policy GKD~\citep{agarwal2024onpolicy}, OPD~\citep{lu2025onpolicydistillation}, OPSD~\citep{zhao2026selfdistilledreasoneronpolicyselfdistillation}, RLSD~\citep{yang2026selfdistilledrlvr}, RLCSD~\citep{pan2026rlcsdreinforcementlearningcontrastive}, and VAttnPO. Each method retains its original objective while sharing the training data, optimization budget, and evaluation protocol. GRPO and VAttnPO additionally share the verifiable reward, sampling, clipping, and KL regularization; their sole method difference is that VAttnPO replaces the broadcast sequence advantage with Equation~(8). Appendices~\ref{app:data} and \ref{app:repro} provide the full data composition, reward, sampling, and method parameters.\looseness=-1

Table~\ref{tab:main} reports VStar~\citep{wu2024v}, HRBench~\citep{wang2025divide}, MMVP~\citep{tong2024eyes}, MMMU-Pro~\citep{mmmupro}, MuirBench~\citep{muirbench}, and BLINK~\citep{blink}, whose mean is AVE. These tasks span general visual reasoning, multidisciplinary understanding, multi-image relations, fine-grained perception, and visual pattern discrimination, so AVE summarizes complementary image capabilities rather than mixing every metric into one scale. Table~\ref{tab:structured} separately reports ChartQA~\citep{chartqa}, OCRBench~\citep{Liu_2024}, DocVQA~\citep{mathew2021docvqa}, InfoVQA~\citep{mathew2022infographicvqa}, POPE~\citep{pope}, HallusionBench~\citep{hallusionbench}, and CountBenchQA~\citep{beyer2024paligemma}. Table~\ref{tab:video} reports MVBench~\citep{mvbench}, LongVideoBench~\citep{longvideobench}, TempCompass~\citep{tempcompass}, Video-MME~\citep{videomme}, and MME~\citep{mme}. ChartQA, all video metrics, MME, and the heterogeneously scaled metrics in Table~\ref{tab:structured} do not enter AVE.

\subsection{Main Results}
\begin{table*}[t]
\caption{Image reasoning and fine-grained perception. AVE is the mean of the six tasks shown. Best results are shown in \textbf{bold}.}
\label{tab:main}
\centering
\small
\resizebox{\linewidth}{!}{%
\begin{tabular}{llccccccc}
\toprule
\multirow{2}{*}{Model Type} & \multirow{2}{*}{Method} & \multicolumn{3}{c}{Fine-grained Perception} & \multicolumn{3}{c}{Reasoning} & \multirow{2}{*}{AVE}\\
\cmidrule(lr){3-5}\cmidrule(lr){6-8}
& & VStar & HRBench 4K & MMVP & MMMU-Pro & MuirBench & BLINK & \\
\midrule
Base & Base model & 81.15 & 77.38 & 80.00 & 52.95 & 60.35 & 67.18 & 69.84\\
\midrule
\multirow{2}{*}{RLVR} & GRPO & 83.25 & 77.50 & 81.00 & 54.28 & 61.54 & 68.23 & 70.97\\
& SAPO & 84.29 & 77.75 & 81.33 & 53.82 & 60.27 & 68.33 & 70.97\\
\midrule
\multirow{3}{*}{\shortstack[l]{On-policy\\Distillation}} & GKD & 79.06 & 79.00 & 76.33 & 53.87 & 62.62 & 63.97 & 69.14\\
& OPD & 85.34 & 78.75 & 79.33 & 54.57 & 63.65 & 65.97 & 71.27\\
& OPSD & 82.72 & 80.13 & 80.67 & 54.74 & 60.46 & 66.65 & 70.90\\
\midrule
\multirow{2}{*}{\shortstack[l]{Self-distilled\\RL}} & RLSD & 82.20 & 77.62 & 82.00 & 53.18 & 60.58 & 65.65 & 70.21\\
& RLCSD & 83.77 & 77.50 & 79.67 & 54.10 & 61.85 & 69.02 & 70.99\\
\midrule
\rowcolor{gray!15}\textbf{Ours} & \best{VAttnPO} & \best{87.43} & \best{80.62} & \best{82.33} & \best{55.95} & \best{65.62} & \best{69.07} & \best{73.50}\\
\bottomrule
\end{tabular}}
\end{table*}
\paragraph{Image Reasoning and Fine-Frained Perception.}
VAttnPO raises AVE from 70.97 with GRPO to 73.50 ($+2.53$) and exceeds the strongest compared baseline, OPD at 71.27, by 2.23 points. All six image tasks improve over GRPO: the largest gains are 4.18 on VStar and 4.08 on MuirBench, while the other four tasks also remain positive. This distribution shows that the average gain spans both fine-grained perception and visual reasoning rather than being driven by one benchmark.

\begin{table*}[t]
\caption{Structured, document, and reliability results. HallusionBench reports aAcc only. Best results are shown in \textbf{bold}.}
\label{tab:structured}
\centering
\small
\resizebox{\linewidth}{!}{%
\begin{tabular}{llccccccc}
\toprule
\multirow{2}{*}{Model Type} & \multirow{2}{*}{Method} & \multicolumn{4}{c}{Structured \& Document} & \multicolumn{2}{c}{Hallucination} & \multirow{2}{*}{CountBenchQA}\\
\cmidrule(lr){3-6}\cmidrule(lr){7-8}
& & ChartQA & OCRBench & DocVQA & InfoVQA & POPE & HallusionBench & \\
\midrule
Base & Base model & 82.72 & 88.10 & 95.27 & 82.38 & 87.77 & 70.87 & 88.71\\
\midrule
\multirow{2}{*}{RLVR} & GRPO & 82.92 & 89.50 & 95.40 & 82.97 & 88.21 & 71.61 & 89.53\\
& SAPO & 82.92 & 89.50 & 95.41 & 82.44 & 88.19 & 70.56 & 89.94\\
\midrule
\multirow{3}{*}{\shortstack[l]{On-policy\\Distillation}} & GKD & 81.36 & 88.70 & 94.86 & 82.11 & 88.44 & 71.50 & 85.42\\
& OPD & 81.44 & 88.70 & 95.04 & 82.18 & 87.58 & 73.29 & 88.71\\
& OPSD & 82.60 & 90.00 & 95.26 & 82.38 & 88.44 & 71.19 & 89.73\\
\midrule
\multirow{2}{*}{\shortstack[l]{Self-distilled\\RL}} & RLSD & 81.44 & 89.50 & 94.87 & 81.89 & 88.39 & 70.14 & 89.12\\
& RLCSD & 83.20 & 89.40 & 95.24 & 82.85 & 87.92 & 72.03 & 89.12\\
\midrule
\rowcolor{gray!15}\textbf{Ours} & \best{VAttnPO} & \best{83.84} & \best{90.10} & \best{95.59} & \best{83.64} & \best{88.67} & \best{73.82} & \best{91.17}\\
\bottomrule
\end{tabular}}
\end{table*}
\paragraph{Structured, Document, and Reliability.}
VAttnPO obtains the highest score among compared methods on all seven structured, document, and reliability metrics in Table~\ref{tab:structured}. Its largest margins over GRPO are 2.21 points on HallusionBench aAcc and 1.64 on CountBenchQA~\citep{beyer2024paligemma}, whereas gains on document tasks are smaller, including 0.19 on DocVQA. The magnitudes are therefore heterogeneous, but their direction extends the improvement beyond the six image tasks to visual reliability, counting, and document understanding.

\begin{table*}[t]
\caption{Video understanding and MME. MME-per and MME-rea denote MME-Perception and MME-Reasoning. Best results are shown in \textbf{bold}.}
\label{tab:video}
\centering
\small
\resizebox{\linewidth}{!}{%
\begin{tabular}{llcccccc}
\toprule
\multirow{2}{*}{Model Type} & \multirow{2}{*}{Method} & \multicolumn{4}{c}{Video Understanding} & \multicolumn{2}{c}{MME}\\
\cmidrule(lr){3-6}\cmidrule(lr){7-8}
& & MVBench & LongVideoBench & TempCompass & Video-MME & MME-per & MME-rea\\
\midrule
Base & Base model & 70.72 & 62.2 & 69.10 & 67.7 & 1724.7 & 671.4\\
\midrule
\multirow{2}{*}{RLVR} & GRPO & 71.25 & 62.8 & 70.04 & 68.4 & 1736.1 & 673.9\\
& SAPO & 71.58 & 62.9 & 69.97 & 68.4 & 1733.2 & 652.1\\
\midrule
\multirow{3}{*}{\shortstack[l]{On-policy\\Distillation}} & GKD & 68.53 & 57.3 & 70.04 & 64.4 & 1722.7 & 656.8\\
& OPD & 70.97 & 62.1 & 71.38 & 66.6 & 1727.5 & 681.8\\
& OPSD & 70.83 & 62.7 & 70.89 & 67.6 & 1732.4 & 657.9\\
\midrule
\multirow{2}{*}{\shortstack[l]{Self-distilled\\RL}} & RLSD & 70.08 & 61.0 & 68.77 & 66.4 & 1724.6 & 658.9\\
& RLCSD & 70.60 & 62.6 & 70.46 & 68.0 & 1732.3 & 661.1\\
\midrule
\rowcolor{gray!15}\textbf{Ours} & \best{VAttnPO} & \best{72.15} & \best{63.6} & \best{71.58} & \best{68.6} & \best{1755.0} & \best{683.9}\\
\bottomrule
\end{tabular}}
\end{table*}
\paragraph{Video Understanding and MME.} VAttnPO's gains extend beyond the six image tasks. It improves all four video benchmarks over GRPO, by 0.90 on MVBench, 0.8 on LongVideoBench, and 1.54 on TempCompass, with a smaller margin of 0.2 on Video-MME. It also improves MME-per and MME-rea by 18.9 and 10.0 points, respectively (Table~\ref{tab:video}). 
Consequently, these performance improvements span short-form video, long-form video, temporal aspects, and other multimodal evaluations, with the magnitude of improvement varying across different benchmarks.

\subsection{Ablation and Controlled Analysis}
\label{sec:ablation}
\begin{wrapfigure}{r}{0.60\linewidth}
\vspace{-0.6\baselineskip}
\centering
\begin{minipage}[t]{0.49\linewidth}
\centering
\includegraphics[width=\linewidth]{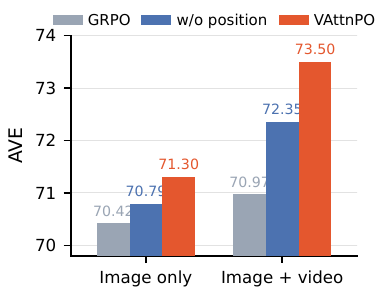}\\[-1mm]
\small (a) Data composition
\end{minipage}\hfill
\begin{minipage}[t]{0.49\linewidth}
\centering
\includegraphics[width=\linewidth]{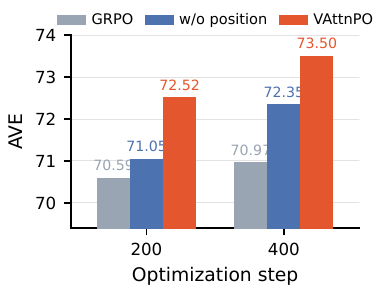}\\[-1mm]
\small (b) Checkpoint comparison
\end{minipage}
\caption{(a) AVE under image-only and image--video training. (b) AVE at steps 200 and 400.}
\label{fig:ablation-plots}
\vspace{-0.6\baselineskip}
\end{wrapfigure}

We examine VAttnPO through controls on its components and similarity metric, position design, reference and token correspondence, and credit balancing. Figure~\ref{fig:ablation-plots} additionally compares data compositions and training checkpoints. Every AVE uses the six tasks in Table~\ref{tab:main}.

\begin{table*}[t]
\centering
\caption{Four controlled ablation groups. AVE denotes the mean over the six image tasks in Table~\ref{tab:main}.}
\label{tab:ablations}
\begingroup
\footnotesize
\captionsetup[subtable]{font=small,skip=3pt,justification=centering}
\setlength{\tabcolsep}{3pt}
\renewcommand{\arraystretch}{1.05}

\begin{subtable}[t]{0.48\textwidth}
\centering
\caption{Core components and metric}
\label{tab:components}
\begin{tabularx}{\linewidth}{@{}Ycccc@{}}
\toprule
Variant & Metric & Position & Balance & AVE\\
\midrule
GRPO & None & No & No & 70.97\\
Position-only & None & Yes & Yes & 70.84\\
KL variant & KL & Yes & Yes & 71.37\\
No position & JSD & No & Yes & 72.35\\
\best{VAttnPO} & \best{JSD} & \best{Yes} & \best{Yes} & \best{73.50}\\
\bottomrule
\end{tabularx}
\end{subtable}\hfill
\begin{subtable}[t]{0.48\textwidth}
\centering
\caption{Position design}
\label{tab:position}
\begin{tabularx}{\linewidth}{@{}YYc@{}}
\toprule
Variant & Setting & AVE\\
\midrule
GRPO & No visual credit & 70.97\\
Position-only & Pure position & 70.84\\
No position & Uniform & 72.35\\
Position-shuffled & Shuffled sigmoid & 72.19\\
\best{VAttnPO} & \best{Relative sigmoid} & \best{73.50}\\
\bottomrule
\end{tabularx}
\end{subtable}

\par\medskip
\begin{subtable}[t]{0.48\textwidth}
\centering
\caption{Reference and correspondence}
\label{tab:reference-token}
\begin{tabularx}{\linewidth}{@{}Yccc@{}}
\toprule
Variant & Reference & Token credit & AVE\\
\midrule
GRPO & None & N/A & 70.97\\
Wrong ref. & Wrong GT & Aligned & 68.92\\
Shuffled ref. & Shuffled GT & Aligned & 70.46\\
Token-permuted & GT & Permuted & 70.86\\
\best{VAttnPO} & \best{GT} & \best{Aligned} & \best{73.50}\\
\bottomrule
\end{tabularx}
\end{subtable}\hfill
\begin{subtable}[t]{0.48\textwidth}
\centering
\caption{Credit balance}
\label{tab:balance}
\begin{tabularx}{\linewidth}{@{}Yccc@{}}
\toprule
Variant & \shortstack{Guarantees\\zero sum} & \shortstack{Preserves\\credit sign} & AVE\\
\midrule
GRPO & N/A & N/A & 70.97\\
No balance & No & Yes & 70.84\\
Mean recenter & Yes & No & 71.35\\
\best{VAttnPO} & \best{Yes} & \best{Yes} & \best{73.50}\\
\bottomrule
\end{tabularx}
\end{subtable}
\endgroup
\end{table*}

\paragraph{Core components and metric.}
Position-only replaces visual credit with a centered, RMS-calibrated position curve; No position retains visual credit with uniform position weights; and the KL variant replaces JSD similarity with reference smoothing, forward KL, and an exponential map while retaining position modulation and credit balancing. Position-only reaches 70.84 AVE, close to GRPO at 70.97; No position reaches 72.35, and the full VAttnPO reaches 73.50. Visual credit therefore contributes beyond a positional prior, with position modulation providing a further gain. The KL variant reaches 71.37; this comparison concerns only the construction defined in Appendix~\ref{app:controls} and leaves policy KL regularization unchanged.

\paragraph{Position Design.}
Position-shuffled permutes the same position weights within each response, preserving their values while breaking their correspondence with tokens. Relative-position weighting raises AVE from 72.35 for No position to 73.50, whereas Position-shuffled reaches 72.19. Together with Position-only at 70.84, this pattern favors coupling visual credit to its original relative positions.

\paragraph{Reference And Token Correspondence.}
Wrong-GT reference replaces the teacher-forced correct option with a deterministic incorrect option; Shuffled reference permutes only the reference's visual-patch indices; and Token-permuted shuffles centered credits $r_{i,t}$ within each response. They reach 68.92, 70.46, and 70.86 AVE, respectively, compared with 73.50 for the complete VAttnPO, showing that reference content, visual correspondence, and token correspondence each affect the construction. 

\paragraph{Credit Balancing.}
No balance directly uses the position-weighted credit $u_{i,t}$; Mean recenter restores zero sum by subtracting the trajectory mean; and the full rule rescales positive and negative credits separately to retain local signs. They reach 70.84, 71.35, and 73.50 AVE, respectively, supporting the combination of zero-sum restoration and local sign preservation.

\paragraph{Data Composition.}
VAttnPO improves over GRPO with both image-only and image--video training, with a larger gain on the mixed data: $+0.88$ and $+2.53$ AVE, respectively (Figure~\ref{fig:ablation-plots}(a)). The variant without position weighting lies between GRPO and the full method in both settings.

\paragraph{Checkpoint Comparison.}
VAttnPO leads GRPO and the variant without position weighting at both evaluated checkpoints, reaching 72.52 and 73.50 AVE at steps 200 and 400, respectively (Figure~\ref{fig:ablation-plots}(b)). This comparison concerns two discrete training checkpoints rather than a continuous convergence trend.

\subsection{Cross-Backbone Generalization}
\begin{wraptable}{r}{0.55\linewidth}
\vspace{-0.7\baselineskip}
\centering
\caption{Cross-backbone AVE. $\Delta$ denotes the VAttnPO gain over GRPO.}
\label{tab:cross-backbone-main}
\footnotesize
\setlength{\tabcolsep}{3pt}
\begin{tabularx}{\linewidth}{@{}Ycccc@{}}
\toprule
Backbone & Base & GRPO & VAttnPO & $\Delta$\\
\midrule
Qwen2.5-VL-7B-Instruct & 59.44 & 61.64 & \best{62.55} & \best{+0.91}\\
InternVL3-8B-Instruct & 59.54 & 60.46 & \best{61.03} & \best{+0.56}\\
\bottomrule
\end{tabularx}
\vspace{-0.5\baselineskip}
\end{wraptable}
VAttnPO continues to demonstrate performance gains over GRPO across two additional backbone models. Specifically, VAttnPO improves the AVE by 0.91 on Qwen2.5-VL-7B~\citep{bai2025qwen2} and by 0.56 on InternVL3-8B~\citep{internvl3exploringadvancedtraining}. The detail results for each benchmark are listed in Appendix~\ref{app:crossbackbone}.

\section{Conclusion}
We introduced VAttnPO, which converts answer-conditioned visual attention from a frozen old policy into a zero-mean, position-aware correction to the GRPO sequence advantage. On Qwen3-VL-8B-Instruct, VAttnPO raises six-task image AVE from 70.97 to 73.50, scores higher than GRPO on every reported structured, video, and MME metric, and retains positive gains on two additional backbones. Together, the diagnostics and controlled comparisons show that internal visual statistics can refine within-response credit while preserving the trajectory-mean advantage.

\subsection*{AI use statement}
We used generative AI tools to polish the writing, check mathematical derivations and formulas, and assist with writing and running parts of the research code, including submitting training jobs. The authors reviewed the edited text, verified the mathematical arguments, and inspected the AI-assisted code and resulting runs. We take full responsibility for the final manuscript and results.

\section*{Reproducibility Statement}
    We document the training-data filtering and composition in Appendix~\ref{app:data}, and report optimization settings, sampling parameters, random seeds, and the software and hardware environment in Appendix~\ref{app:repro}. Section~4 and 
    Appendix~\ref{app:implementation} specify the visual-reference construction, attention aggregation, numerical handling, and policy objective, with the update procedure summarized in Algorithm~\ref{alg:vattnpo}. Derivations of the credit-balancing and 
    trajectory-mean preservation properties appear in Appendix~\ref{app:properties}. Diagnostic protocols, statistical procedures, controlled variants, and benchmark evaluation are described in Appendices~\ref{app:diagnostics}--\ref{app:evaluation}. 
    Appendices~\ref{app:crossbackbone} and \ref{app:layerwise} provide additional cross-backbone results and layer-wise analyses.

\bibliography{references}
\bibliographystyle{iclr2027_conference}

\clearpage
\appendix
\numberwithin{figure}{section}
\numberwithin{table}{section}

\section{Training Data and Preprocessing}
\label{app:data}
\subsection{Data Sources and Composition}
The main training data are cleaned multiple-choice samples selected from Video-R1-260k~\citep{Video-R1}. All primary methods use the same frozen training manifest. The data-composition ablation changes only this manifest and leaves the other training settings unchanged.
\begin{table}[h]
\centering
\caption{Training-set composition.}
\begin{tabular}{lr}
\toprule
Data type & Number of examples\\
\midrule
Image & 64,473\\
Video & 103,987\\
Total & 168,460\\
\bottomrule
\end{tabular}
\end{table}

\subsection{Filtering and Preprocessing}
Starting from 263,071 examples in Video-R1-260k, we retain the 168,769 examples whose \texttt{problem\_type} is multiple choice and whose \texttt{data\_type} is image or video. The answer letter is extracted from the \texttt{<answer>...</answer>} field in the solution, and 10 examples without an extractable answer letter are removed. We then reserve the final 32 examples in the original data order for validation, leaving 168,727 candidate training examples.

An independent video-availability scan reads the first and last frames of 26,896 unique video paths and identifies 15 invalid paths. After removing the 250 examples associated with these paths, we exclude 17 examples for which no valid incorrect answer can be constructed from the parsed options. The resulting fixed training manifest contains 168,460 examples: 64,473 image examples and 103,987 video examples. The 32 validation examples are not included in this training set.

\section{Reproducibility and Training Configuration}
\label{app:repro}
\subsection{Optimization and Sampling}
GRPO and VAttnPO share the training data, learning rate, random seed, 400-step budget, sampling, and optimization settings. VAttnPO replaces the broadcast sequence advantage with the token advantage constructed in Equation~(8). Each input has $G=8$ sampled responses.
\begin{table}[h]
\centering
\caption{Shared training configuration for GRPO and VAttnPO.}
\begin{tabular}{ll}
\toprule
Item & Setting\\
\midrule
Backbone & Qwen3-VL-8B-Instruct\\
SFT warm-up & None\\
Training steps & 400\\
Loss type & GRPO\\
Learning rate & $2\times10^{-6}$\\
Optimizer / schedule & AdamW Torch Fused / cosine\\
Per-device batch / processes / accumulation & 1 / 8 / 32\\
Nominal global batch size & 256\\
Response group size $G$ & 8\\
Temperature / top-$p$ / top-$k$ & 1.0 / 0.85 / unrestricted\\
Maximum sequence / completion length & 8192 / 4096\\
Dynamic sampling / resampling attempts & Enabled / 3\\
Overlong filtering & Enabled\\
Clipping $\varepsilon$ / ratio range & 0.20 / $[0.80,1.20]$\\
Policy KL coefficient $\beta_{\mathrm{KL}}$ & 0.04\\
Training / data seed & 42 / 42\\
Parameter updates & Full fine-tuning\\
Warm-up ratio & 0.05\\
Precision / distributed strategy & bfloat16 / ZeRO-2\\
\bottomrule
\end{tabular}
\end{table}
VAttnPO constructs visual attention from all 32 query heads at zero-based layer 20 and uses the dataset answer as the reference. We use JSD similarity with $\alpha=1$, $\mu=0.5$, $\kappa=8$, and $g_{\min}=0.05$, followed by position weighting and credit balancing for both image and video responses. This layer is the 21st of 36 decoder blocks; grouped-query attention expands its eight key--value heads to 32 query heads.

\subsection{Verifiable Reward}
The sequence reward for multiple-choice training combines accuracy, format, and length terms:
\[
R=0.9R_{\mathrm{acc}}+0.1R_{\mathrm{fmt}}+0.1R_{\mathrm{over}}.
\]
The accuracy reward extracts an answer from \texttt{<answer>...</answer>}, the final \texttt{\textbackslash boxed\{...\}} expression, or the full response, in that order. If both prediction and ground truth resolve to option letters A--J, they are compared case-insensitively; otherwise, the scorer uses mathematical equivalence. A match gives $R_{\mathrm{acc}}=1$ and a mismatch gives 0.
The format reward checks whether the response ends with the case-sensitive form \texttt{<answer>\textbackslash boxed\{...\}</answer>}; a match gives $R_{\mathrm{fmt}}=1$ and otherwise 0.
Let $L$ denote the completion length in tokens. The length penalty starts at $L_{\mathrm{start}}=3072$ and reaches $-1$ at the maximum completion length of 4096:
\[
R_{\mathrm{over}}(L)=
\min\!\left(-\frac{L-3072}{1024},0\right)
=
\begin{cases}
0, & L\leq3072,\\[1mm]
-\dfrac{L-3072}{1024}, & 3072<L\leq4096.
\end{cases}
\]

\subsection{Software and Hardware}
Training uses ms-swift 4.4.1 with its default visual preprocessing and video-sampling settings, running on eight NVIDIA L20X GPUs with bfloat16 precision and DeepSpeed ZeRO-2. Except for the explicit checkpoint comparison, the main results use the prespecified 400-step endpoint rather than selecting checkpoints by test performance.

\section{VAttnPO Implementation Details}
\label{app:implementation}
\subsection{Visual Reference from the Correct Answer}
For each input, the frozen old policy enters an assistant turn after the original system and multimodal user prompt and teacher-forces the dataset answer. All 168,460 training targets strictly use \texttt{<answer>X</answer>}, where $X\in\{\mathrm{A},\ldots,\mathrm{J}\}$ is the correct option letter. They contain no explanation, chain of thought, or \texttt{\textbackslash boxed\{\}}, and the reference branch appends neither EOS nor \texttt{<|im\_end|>}.
Valid ground-truth queries are defined by \texttt{labels != -100} within the assistant response. The tokens pooled into $P_{\mathrm{gt}}$ therefore include the opening tag, correct option letter, and closing tag, but exclude the assistant header and padding. Their attention to visual keys is averaged over the $L_{\mathrm{gt}}$ queries and all $H$ heads and then normalized over the $N$ visual positions. This branch is stop-gradient; the correct answer supplies conditioning context but no region annotation.

\subsection{Sampled Response Attention}
The main method uses the attention query at the same index as $y_{i,t}$ and constructs retrospective credit after the complete response has been generated. The behavioral intervention in Section~\ref{sec:pilot} instead uses the query immediately before the predicted target token. These indexing choices correspond to different analyses and should not be interchanged. At the target layer, attention softmax is first computed over all causal keys, after which only key columns whose input token IDs denote image or video patches are selected. Text, system/user/assistant, visual-boundary, and padding tokens are excluded.
Selected patches preserve their original order in the model input sequence. Video-frame patches are flattened in the processor's spatiotemporal order. We do not normalize individual frames separately: all frames of a video, and all visual patches in multi-image or multi-video inputs, form one shared support $\mathcal V$. The selected post-softmax weights are renormalized in L1:
\[
Q_{i,t,p}=\frac{a_{i,t,p}}{\sum_{p'\in\mathcal V}a_{i,t,p'}+10^{-8}},
\qquad p\in\mathcal V.
\]
The same visual mask, patch order, and normalization are used for $P_{\mathrm{gt}}$. Both distributions therefore describe spatial attention allocation over one common visual support rather than the total attention weight assigned to visual inputs. Main training uses zero-based layer 20 and averages all 32 query heads after grouped-query expansion, without single-head, top-$k$-head, or other head selection.

\subsection{Masks, Length, and Numerical Boundaries}
Let $L_i$ denote the number of valid completion tokens in a response. The terminal token in the response contributes to $L_i$ and follows the same-index attention-credit rule as the other valid tokens.
Relative position is defined only over these valid tokens. For $L_i\geq2$, we use $p_{i,t}=(t-1)/(L_i-1)$; when $L_i=1$, the position weight is one.
Model computation and the query/key projections use BF16, and the attention-logit matrix multiplication follows the model precision. Logits are explicitly cast to FP32 before softmax; head averaging, patch L1 normalization, JSD, position weighting, balancing, and final credits are then computed in FP32. Both the L1 denominator and logarithms use an epsilon of $10^{-8}$. With natural logarithms, let $M=(P+Q)/2$ and
\[
\operatorname{JSD}(Q\Vert P)
=\frac12\operatorname{KL}(Q\Vert M)
+\frac12\operatorname{KL}(P\Vert M).
\]
The raw JSD lies in $[0,\ln2]$, and the implementation clips $1-\operatorname{JSD}/\ln2$ to $[0,1]$. The KL metric variant's $\epsilon_{\mathrm{KL}}=10^{-4}$ is not used by the main JSD construction. All visual statistics are stop-gradient. Credit balancing uses $\tau=10^{-8}$; if $P_i^+\leq\tau$ or $N_i^-\leq\tau$, the entire $\mathbf d_i$ is set to zero. The fail-closed zero-sum check uses tolerance $10^{-6}\max(\sum_t|d_{i,t}|,1)$.

\subsection{Full Clipped Optimization Objective}
Define the likelihood ratio
\[
\rho_{i,t}(\theta)=
\frac{\pi_\theta(y_{i,t}\mid x,y_{i,<t})}
{\pi_{\theta_{\mathrm{old}}}(y_{i,t}\mid x,y_{i,<t})}.
\]
The frozen old policy $\pi_{\theta_{\mathrm{old}}}$ is used both to sample responses and to extract visual signals; a separate frozen $\pi_{\mathrm{ref}}$ is used only for KL regularization. VAttnPO inserts $\widetilde A_{i,t}$ into the clipped GRPO objective, with symmetric clipping $\varepsilon=0.20$, corresponding to the ratio range $[0.80,1.20]$:
\[
\begin{aligned}
J(\theta)=\mathbb E\Bigg[\frac1{L_i}\sum_t\Big\{
&\min\!\Big(
\rho_{i,t}(\theta)\widetilde A_{i,t},
\operatorname{clip}(\rho_{i,t}(\theta),1-\varepsilon,1+\varepsilon)
\widetilde A_{i,t}
\Big)\\
&-\beta_{\mathrm{KL}}D_{\mathrm{KL},i,t}
\Big\}\Bigg].
\end{aligned}
\]
The position weight acts only on visual credit and never directly scales $\hat A_i$; gradients propagate only through this policy objective.

Algorithm~\ref{alg:vattnpo} summarizes the VAttnPO update.
\begin{algorithm}[t]
\caption{VAttnPO update}
\label{alg:vattnpo}
\begin{algorithmic}[1]
\Require Input $x$, ground-truth answer $y_{\mathrm{gt}}$, rollout policy $\pi_{\theta_{\mathrm{old}}}$, group size $G$, and credit coefficient $\alpha$
\Ensure Updated policy $\pi_\theta$
\State Sample $\{y_i\}_{i=1}^{G}\sim\pi_{\theta_{\mathrm{old}}}$ and compute $\{R_i,\hat A_i\}_{i=1}^{G}$ \Comment{sequence-level signal}
\State Teacher-force $y_{\mathrm{gt}}$ and construct $P_{\mathrm{gt}}$ with Equation~(2) \Comment{shared reference}
\For{$i=1$ to $G$}
  \For{$t=1$ to $L_i$}
    \State Extract $Q_{i,t}$ and compute $s_{i,t}$ with Equation~(3)
  \EndFor
  \State Compute $r_{i,t}$ and $u_{i,t}$ with Equations~(4)--(5)
  \State Compute $d_{i,t}$ with Equations~(6)--(7) \Comment{zero-mean correction}
  \State Set $\widetilde A_{i,t}=\hat A_i+\alpha d_{i,t}$ for all valid $t$
\EndFor
\State Update $\pi_\theta$ with the clipped objective above
\State \Return $\pi_\theta$
\end{algorithmic}
\end{algorithm}

\section{Properties and Derivations}
\label{app:properties}
\subsection{Endpoint-Normalized Position Weight}
For $\kappa>0$ and finite $\mu$, the sigmoid is strictly increasing and the denominator in Equation~(5) is positive. Thus $S_\kappa$ increases strictly from $S_\kappa(0)=0$ to $S_\kappa(1)=1$. With $0<g_{\min}\leq1$, the position weight is positive and nondecreasing on $[0,1]$, with range $[g_{\min},2-g_{\min}]$; it is strictly increasing when $g_{\min}<1$. For $\mu=0.5$, sigmoid symmetry gives $S_\kappa(0.5)=0.5$ and a midpoint weight of one. For $L_i=1$, we use unit weight directly.

\subsection{Centering and Credit Balancing}
Centering removes the shared similarity offset and gives $\sum_t r_{i,t}=0$, but leaves within-trajectory variance unchanged and does not standardize credit scales across responses. After position weighting, the residual satisfies $\sum_tu_{i,t}=L_i\operatorname{Cov}_t(g_i^{\mathrm{pos}},r_i)$, where the covariance is over valid token positions. In a nondegenerate trajectory, Equation~(7) rescales positive and negative credits by positive factors, preserving their signs. The total positive credit after balancing is
\[
\sum_{t:u_{i,t}>0}d_{i,t}=\frac{C_i}{P_i^+}P_i^+=C_i,
\]
and the total negative credit is
\[
\sum_{t:u_{i,t}<0}d_{i,t}=-\frac{C_i}{N_i^-}N_i^-=-C_i.
\]
Their sum is zero, and entries with $u_{i,t}=0$ remain zero. In either degenerate branch, $\mathbf d_i=\mathbf0$, which also has zero sum. This sign property concerns $d_{i,t}$ relative to $u_{i,t}$.
Without position weighting, $u_{i,t}=r_{i,t}$; in exact arithmetic and the nondegenerate case, the two totals are equal, so balancing reduces to the identity $d_{i,t}=r_{i,t}$.

\subsection{Preservation of the Trajectory-Mean Advantage}
Because $\widetilde A_{i,t}=\hat A_i+\alpha d_{i,t}$ and $\sum_t d_{i,t}=0$,
\[
\frac1{L_i}\sum_{t=1}^{L_i}\widetilde A_{i,t}=\hat A_i.
\]
This identity preserves the trajectory mean of advantage coefficients, not token gradients, their vector sum, or their norm. The final advantage can change sign when $\alpha d_{i,t}$ opposes a nonzero $\hat A_i$ and exceeds it in magnitude; when $\hat A_i=0$, a nonzero correction can still produce positive and negative token advantages with zero mean.

\subsection{Computational Complexity}
Equation~(3) computes JSD over an $N$-dimensional visual distribution for each valid rollout token. For $T$ valid rollout tokens in a group, this credit computation requires $O(NT)$ operations.

\section{Diagnostic Protocols and Supplementary Analyses}
\label{app:diagnostics}
\subsection{Shared Sampling and Statistics}
All three diagnostics use layer 20 of Qwen3-VL-8B-Instruct and the same image-only candidate pool. The candidate record contains 10,199 responses to 1,485 prompts from the image distribution of the training set; 732 prompts contain both correct and incorrect responses. We sample 200 of these prompts without replacement using seed 42 and obtain 1,429 valid trajectories, comprising 771 correct and 658 incorrect responses.
For correct--incorrect comparisons, responses are aggregated within each prompt before prompts receive equal weight. The 95\% intervals use 2,000 bootstrap resamples with the prompt as the cluster unit.

\subsection{Analysis 1: Token Intervention}
For each example, select the top 20\% of visual patches under $P_{\mathrm{gt}}$ as $S_{\mathrm{gt}}$. Let $\bar a^{\mathrm{probe}}_{i,t-1,p}$ be head-aggregated attention from the query before $y_{i,t}$ to visual patch $p$. Define
\[
Q^{\mathrm{probe}}_{i,t,p}=
\frac{\bar a^{\mathrm{probe}}_{i,t-1,p}}
{\sum_{p'\in\mathrm{visual}}\bar a^{\mathrm{probe}}_{i,t-1,p'}+\epsilon},
\qquad
w_{i,t}=\sum_{p\in S_{\mathrm{gt}}}Q^{\mathrm{probe}}_{i,t,p}.
\]
The reasoning mask covers reasoning tokens before the committed answer span. For reasoning length $L_r$, every intervention set has size $m=\max(3,\operatorname{round}(0.15L_r))$. High and low select the $m$ tokens with the largest and smallest $w_{i,t}$, respectively. Random samples without replacement from the middle pool after removing high and low tokens. Position-matched greedily assigns each high token the nearest unused non-high token in relative position. The selected tokens' complete self-attention key--value columns are then masked for subsequent queries. The intervention effect is the log-probability drop of the model's original prediction:
\[
\Delta\ell(S)=
\log p_{\mathrm{full}}(y_i^{\mathrm{pred}})-
\log p_{\mathrm{mask}(S)}(y_i^{\mathrm{pred}}).
\]
The paired effects are Cohen's $d=0.957$ (95\% CI $[0.867,1.052]$) for High versus Low, $1.005$ ($[0.925,1.083]$) for High versus Random, and $0.852$ ($[0.764,0.942]$) for High versus Position-matched. The mean relative positions of the High and Position-matched sets are 0.666 and 0.654, respectively; the effect direction is positive in both the correct and incorrect response subsets.
All effect sizes use paired differences within the same prompt and the prompt-cluster bootstrap above; position matching is deterministic and greedy.

\subsection{Analysis 2: GT versus Shuffled Visual Reference}
We compute each sampled token distribution's JSD from the GT reference $P_{\mathrm{gt}}$ and from a patch-index-shuffled version. The permutation preserves the multiset of reference values but breaks visual-patch identity. For an example $e$ containing both correct and incorrect responses, the full-completion paired gap is
\[
\delta_e=
\operatorname{mean}_{i\in\mathcal W_e}\operatorname{mean}_tD_{i,t}-
\operatorname{mean}_{i\in\mathcal C_e}\operatorname{mean}_tD_{i,t},
\]
where $\mathcal W_e$ and $\mathcal C_e$ are its incorrect and correct responses. The comparison statistic is the difference between the GT-reference and shuffled-reference gaps.
One patch permutation is generated and fixed for each $(\texttt{source\_row\_idx},\texttt{rollout\_index})$ pair. All token subsets of the same response share it, and it is not resampled per token or bootstrap replicate. This pilot uses the image-only candidate pool, so no video-frame boundary is involved.

\subsection{Analysis 3: Correct--Incorrect JSD over Completion Progress}
\label{app:diagnostics-progress}
For $L_i>1$, valid tokens in each response are assigned to 20 bins by relative generation position:
\[
b_{i,t}=\left\lfloor19\frac{t-1}{L_i-1}\right\rfloor.
\]
The final bin contains only the terminal token of each response rather than an equal-width interval over its final segment. We first aggregate within each response and bin, then across correct or incorrect responses within each example, and finally weight examples equally.

\section{Ablation and Control Construction}
\label{app:controls}
\subsection{Position and Balance Variants}
\begin{table}[h]
\centering
\caption{Controlled variants and their sole change from VAttnPO.}
\begin{tabularx}{\linewidth}{@{}lY@{}}
\toprule
Variant & Sole change\\
\midrule
Position-only & Center and unit-RMS normalize each trajectory's position curve, then multiply by the fixed $r$-scale proxy 0.14\\
No position & Set $g^{\mathrm{pos}}_{i,t}=1$, so $u_{i,t}=r_{i,t}$\\
Position-shuffled & Deterministically permute $g^{\mathrm{pos}}_{i,t}$ over valid tokens per response with seed 42\\
No balance & Set $d_{i,t}=u_{i,t}$\\
Mean recenter & Set $d_{i,t}=u_{i,t}-L_i^{-1}\sum_vu_{i,v}$\\
\bottomrule
\end{tabularx}
\end{table}
\begin{table}[h]
\centering
\caption{Per-benchmark results for position and credit-balance variants.}
\resizebox{\linewidth}{!}{%
\begin{tabular}{lccccccc}
\toprule
Variant & VStar & MMMU-Pro & MuirBench & BLINK & HRBench 4K & MMVP & AVE\\
\midrule
GRPO & 83.25 & 54.28 & 61.54 & 68.23 & 77.50 & 81.00 & 70.97\\
Position-only & 83.34 & 54.18 & 62.23 & 67.85 & 78.12 & 79.33 & 70.84\\
No balance & 83.77 & 54.41 & 62.04 & 67.75 & 77.38 & 79.67 & 70.84\\
Mean recenter & 84.82 & 54.47 & 62.04 & 68.23 & 77.88 & 80.67 & 71.35\\
No position & 85.34 & 55.55 & 63.35 & 68.70 & 79.50 & 81.67 & 72.35\\
Position-shuffled & 84.95 & 55.48 & 63.25 & 68.39 & 79.38 & 81.67 & 72.19\\
\best{VAttnPO} & \best{87.43} & \best{55.95} & \best{65.62} & \best{69.07} & \best{80.62} & \best{82.33} & \best{73.50}\\
\bottomrule
\end{tabular}}
\end{table}

\paragraph{Similarity metric control.}
The KL variant uses the forward divergence $D_{\mathrm{KL}}(Q_{i,t}\Vert P_{\mathrm{gt}}^{\epsilon})$. We smooth the reference with the uniform distribution $U_j=1/N$, using $\epsilon_{\mathrm{KL}}=10^{-4}$, and map the divergence to similarity with a fixed temperature $\tau_{\mathrm{KL}}=1$:
\[
\begin{aligned}
P_{\mathrm{gt}}^{\epsilon}
&=(1-\epsilon_{\mathrm{KL}})P_{\mathrm{gt}}+\epsilon_{\mathrm{KL}}U,\\
D^{QP}_{i,t}
&=D_{\mathrm{KL}}(Q_{i,t}\Vert P_{\mathrm{gt}}^{\epsilon}),\\
s^{\mathrm{KL}}_{i,t}
&=\exp\!\left(-D^{QP}_{i,t}/\tau_{\mathrm{KL}}\right),\\
r^{\mathrm{KL}}_{i,t}
&=s^{\mathrm{KL}}_{i,t}-\frac1{L_i}\sum_{v=1}^{L_i}s^{\mathrm{KL}}_{i,v}.
\end{aligned}
\]
Centering is performed over valid completion tokens. The same position weights then form $u^{\mathrm{KL}}_{i,t}=g^{\mathrm{pos}}_{i,t}r^{\mathrm{KL}}_{i,t}$, followed by the same credit-balancing rule to obtain $d^{\mathrm{KL}}_{i,t}$. This replaces the similarity construction in Stage 2; policy KL regularization remains unchanged.
The variant scores 84.82, 54.16, 60.92, 68.12, 77.88, and 82.33 on VStar, MMMU-Pro, MuirBench, BLINK, HRBench 4K, and MMVP, respectively, for an AVE of 71.37. The exponential map changes the scale of within-trajectory similarity differences, so this comparison concerns the specified KL construction.

\subsection{Reference and Token-Correspondence Controls}
Shuffled reference applies a fixed permutation to the visual-patch indices, $P_{\mathrm{gt}}^\pi[j]=P_{\mathrm{gt}}[\pi(j)]$, preserving the multiset of probabilities, entropy, and concentration while disrupting spatial correspondence. Image and video patches are permuted separately, without crossing modalities; patches within each modality share one support, allowing permutations across images or video frames. Each example in the current training manifest contains either one image or one video. The permutation is determined per example with seed 42, shared by all responses to that input, and fixed throughout training.
Token-permuted directly permutes the normalized centered similarity credit $r_{i,t}=s_{i,t}-\bar s_i$ from Equation~(4) among valid completion positions, where $s_{i,t}=1-\operatorname{JSD}(Q_{i,t}\Vert P_{\mathrm{gt}})/\ln2$. Let $\pi_i$ denote the permutation for response $i$; the position weights remain at their original positions:
\[
u^{\mathrm{perm}}_{i,t}=g^{\mathrm{pos}}_{i,t}r_{i,\pi_i(t)},
\qquad
\mathbf d_i^{\mathrm{perm}}=\operatorname{Balance}(\mathbf u_i^{\mathrm{perm}}),
\]
where $\operatorname{Balance}$ follows Equations~(6)--(7) and the same degenerate-case rule. With seed 42, the permutation is determined by the input and complete response-token sequence, so different responses generally receive different mappings while identical responses receive the same mapping. It is not resampled by training step, and padding positions are excluded. The operation preserves the multiset, mean, and RMS of $\mathbf r_i$, but need not preserve the multiset or RMS of $\mathbf u_i$ or $\mathbf d_i$ after position weighting and balancing; the balanced credit still has zero sum.

The Wrong-GT reference parses valid options, orders them alphabetically, and selects the first option different from the ground truth. It teacher-forces \texttt{<answer>X</answer>} without \texttt{\textbackslash boxed\{\}}, where $X$ is the selected incorrect letter; its training manifest and all other settings match VAttnPO.

\section{Evaluation Protocol}
\label{app:evaluation}
All benchmark evaluations use VLMEvalKit~\citep{duan2026vlmevalkitopensourcetoolkitevaluating} and follow its default protocol for each benchmark, including prompting, decoding, answer extraction, and scoring. We apply no benchmark-specific protocol overrides.
AVE is the equally weighted mean of VStar, HRBench 4K, MMVP, MMMU-Pro, MuirBench, and BLINK. ChartQA, all video metrics, MME, and the heterogeneously scaled metrics in Table~\ref{tab:structured} are excluded.

\section{Cross-Backbone Results}
\label{app:crossbackbone}
The VAttnPO runs for Qwen2.5-VL-7B-Instruct and InternVL3-8B-Instruct both use zero-based layer 14. Apart from the backbone and attention layer, their settings follow Table~B.1.
\begin{table}[h]
\centering
\caption{Cross-backbone results on the six tasks used to compute AVE.}
\resizebox{\linewidth}{!}{%
\begin{tabular}{llccccccc}
\toprule
Backbone & Method & VStar & MMMU-Pro & MuirBench & BLINK & HRBench 4K & MMVP & AVE\\
\midrule
\multirow{3}{*}{Qwen2.5-VL-7B-Instruct}
& Base & 73.82 & 38.44 & 53.42 & 51.76 & 68.88 & 70.33 & 59.44\\
& GRPO & 75.92 & 39.08 & 53.69 & 54.55 & 72.25 & 74.33 & 61.64\\
& \best{VAttnPO} & \best{76.39} & \best{40.02} & \best{55.58} & \best{54.71} & \best{73.62} & \best{75.00} & \best{62.55}\\
\midrule
\multirow{3}{*}{InternVL3-8B-Instruct}
& Base & 68.59 & 26.36 & 55.65 & 56.60 & 70.37 & 79.67 & 59.54\\
& GRPO & 71.20 & 27.59 & 56.42 & 56.81 & 70.75 & 80.00 & 60.46\\
& \best{VAttnPO} & \best{71.68} & \best{28.36} & \best{56.98} & \best{57.08} & \best{71.38} & \best{80.67} & \best{61.03}\\
\bottomrule
\end{tabular}}
\end{table}

\section{Layer-Wise Visual Attention Analysis}
\label{app:layerwise}

\subsection{Layer-Wise Separation During Inference}
\label{app:inference-layer-localization}
We compare correctness-related visual-attention separation across network layers during inference. All three frozen models use the same fixed set of 200 diagnostic prompts as Section~\ref{sec:pilot}, with $G=8$ sampled responses per prompt, temperature 1.0, top-$p$ 1.0, and generation seed 271828. Each backbone contributes only prompts containing both valid correct and incorrect responses, yielding 198, 44, and 173 paired prompts, respectively; the eligible subsets can therefore differ across backbones.

We use the 0-based completion-progress bins defined in Appendix~\ref{app:diagnostics-progress}. At each layer, visual attention is first aggregated across heads as in Equation~(2), after which JSD is computed. For layer $\ell$, prompt $p$, and progress bin $b$, the within-prompt wrong-minus-correct gap is
\[
\delta^{(\ell)}_{p,b}
=
\operatorname{mean}_{i\in\mathcal W_p}
\operatorname{mean}_{t\in b(i)}D^{(\ell)}_{i,t}
-
\operatorname{mean}_{i\in\mathcal C_p}
\operatorname{mean}_{t\in b(i)}D^{(\ell)}_{i,t},
\]
where $\mathcal W_p$ and $\mathcal C_p$ denote the incorrect and correct sampled responses. We summarize late-completion separation by
\[
\Delta^{\mathrm{late}}_{\ell}
=
\frac{1}{5}\sum_{b=15}^{19}
\operatorname{mean}_{p}\delta^{(\ell)}_{p,b}.
\]
This statistic gives equal weight to the final five bins, including the terminal-token bin, and covers approximately 78.9--100\% completion progress; it is not a pooled average over all tokens in that span. A positive value indicates that incorrect responses are farther from the answer-conditioned visual-attention reference. Confidence intervals use 2,000 prompt-cluster bootstrap resamples. For zero-based layer index $\ell$ in a model with $L_{\mathrm{dec}}$ decoder layers, normalized depth is $\ell/(L_{\mathrm{dec}}-1)$.

\begin{table*}[t]
\centering
\caption{Layer-wise wrong-minus-correct JSD gaps, averaged equally over the final five completion-progress bins. Positive gaps indicate lower reference alignment for incorrect responses; intervals are 95\% prompt-cluster bootstrap. Bold marks Qwen3-VL layer 20, used in the main experiments, and does not denote statistical superiority.}
\label{tab:inference-late-layer-all}
\begingroup
\scriptsize
\setlength{\tabcolsep}{2.8pt}
\renewcommand{\arraystretch}{1.08}
\begin{tabular}{@{}c@{\hspace{0.01\textwidth}}c@{\hspace{0.01\textwidth}}c@{}}
\begin{minipage}[t]{0.32\textwidth}
\centering
\textbf{(a) Qwen3-VL-8B-Instruct}\par\smallskip
\resizebox{\linewidth}{!}{%
\begin{tabular}{rrrr}
\toprule
Layer & Depth & $\Delta^{\mathrm{late}}_{\ell}$ & 95\% CI\\
\midrule
L8  & 23\% & $4.4\times10^{-3}$ & $[1.8\times10^{-3},\,7.1\times10^{-3}]$\\
L12 & 34\% & $3.6\times10^{-3}$ & $[7.0\times10^{-4},\,6.6\times10^{-3}]$\\
L16 & 46\% & $1.0\times10^{-2}$ & $[6.3\times10^{-3},\,1.4\times10^{-2}]$\\
\textbf{L20} & \textbf{57\%} & $\mathbf{1.1\times10^{-2}}$ & $\mathbf{[6.5\times10^{-3},\,1.5\times10^{-2}]}$\\
L24 & 69\% & $6.7\times10^{-3}$ & $[3.2\times10^{-3},\,1.0\times10^{-2}]$\\
L28 & 80\% & $6.9\times10^{-3}$ & $[3.2\times10^{-3},\,1.1\times10^{-2}]$\\
L32 & 91\% & $5.3\times10^{-3}$ & $[2.4\times10^{-3},\,8.2\times10^{-3}]$\\
\bottomrule
\end{tabular}}
\end{minipage}
&
\begin{minipage}[t]{0.32\textwidth}
\centering
\textbf{(b) Qwen2.5-VL-7B-Instruct}\par\smallskip
\resizebox{\linewidth}{!}{%
\begin{tabular}{rrrr}
\toprule
Layer & Depth & $\Delta^{\mathrm{late}}_{\ell}$ & 95\% CI\\
\midrule
L0  & 0\% & $8.2\times10^{-4}$ & $[9.5\times10^{-5},\,1.6\times10^{-3}]$\\
L4  & 15\% & $3.5\times10^{-4}$ & $[-4.1\times10^{-3},\,4.5\times10^{-3}]$\\
L9  & 33\% & $1.1\times10^{-3}$ & $[-4.1\times10^{-3},\,5.6\times10^{-3}]$\\
L14 & 52\% & $4.8\times10^{-3}$ & $[-1.5\times10^{-3},\,1.1\times10^{-2}]$\\
L18 & 67\% & $3.5\times10^{-3}$ & $[-8.2\times10^{-4},\,7.3\times10^{-3}]$\\
L22 & 81\% & $2.8\times10^{-3}$ & $[-3.3\times10^{-3},\,8.9\times10^{-3}]$\\
L27 & 100\% & $2.8\times10^{-3}$ & $[-2.3\times10^{-3},\,8.1\times10^{-3}]$\\
\bottomrule
\end{tabular}}
\end{minipage}
&
\begin{minipage}[t]{0.32\textwidth}
\centering
\textbf{(c) InternVL3-8B-Instruct}\par\smallskip
\resizebox{\linewidth}{!}{%
\begin{tabular}{rrrr}
\toprule
Layer & Depth & $\Delta^{\mathrm{late}}_{\ell}$ & 95\% CI\\
\midrule
L0  & 0\% & $6.9\times10^{-4}$ & $[1.1\times10^{-4},\,1.3\times10^{-3}]$\\
L4  & 15\% & $1.3\times10^{-3}$ & $[-1.1\times10^{-3},\,3.3\times10^{-3}]$\\
L9  & 33\% & $2.8\times10^{-3}$ & $[-3.2\times10^{-4},\,5.5\times10^{-3}]$\\
L14 & 52\% & $4.3\times10^{-3}$ & $[-2.0\times10^{-3},\,1.1\times10^{-2}]$\\
L18 & 67\% & $4.3\times10^{-3}$ & $[1.6\times10^{-4},\,8.7\times10^{-3}]$\\
L22 & 81\% & $3.9\times10^{-3}$ & $[-7.7\times10^{-4},\,8.6\times10^{-3}]$\\
L27 & 100\% & $2.3\times10^{-3}$ & $[-1.6\times10^{-4},\,5.0\times10^{-3}]$\\
\bottomrule
\end{tabular}}
\end{minipage}
\\
\end{tabular}
\endgroup
\end{table*}

Among the scanned layers, the largest point estimates occur at intermediate normalized depths: 0.57 for Qwen3-VL-8B and 0.52 for both Qwen2.5-VL-7B-Instruct and InternVL3-8B-Instruct. In Qwen3-VL-8B, L20 is close to L16: their paired difference is $6.6\times10^{-4}$, with a 95\% bootstrap interval of $[-2.5,3.6]\times10^{-3}$. The Qwen2.5-VL estimates are less precise with 44 paired prompts, while the displayed InternVL3 estimates for L14 and L18 both round to $4.3\times10^{-3}$. The observed profile therefore provides supporting context for using Qwen3-VL layer 20 rather than identifying a uniquely preferred layer.

\subsection{Qualitative Layer-Wise Attention Examples}
\label{app:layerwise-examples}
The quantitative analysis summarizes correctness-related separation across layers; the following examples illustrate spatial attention patterns at individual layers. Figure~\ref{fig:layerwise-examples} presents nine benchmark examples as one continuous visualization. Each PDF shows the original input and head-mean attention heatmaps from layers 8, 12, 16, 20, 24, 28, and 32, with the question and reference answer printed above. Layers within an example use a shared q95 color scale, fixed opacity, and identical smoothing, allowing qualitative inspection of spatial migration across layers within that example; absolute color intensity should not be compared across examples. The figure illustrates qualitative changes in attention patterns across layers rather than establishing causal attribution or correctness.

\begingroup
\centering
\includegraphics[width=\linewidth]{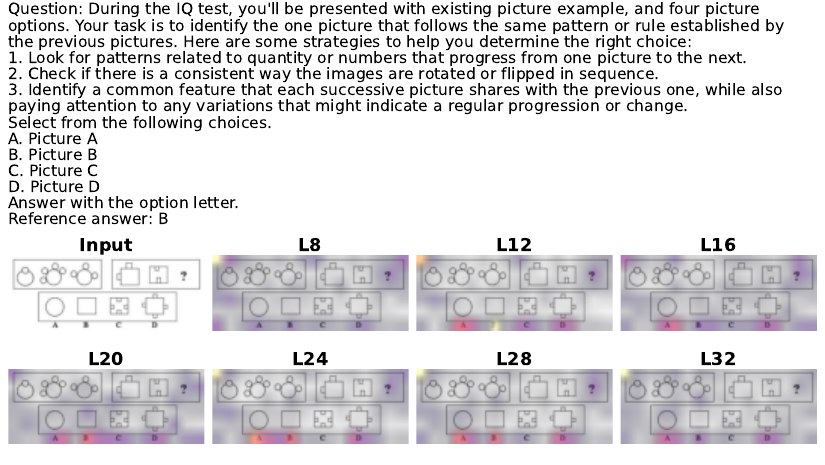}\par\nointerlineskip
\includegraphics[width=\linewidth]{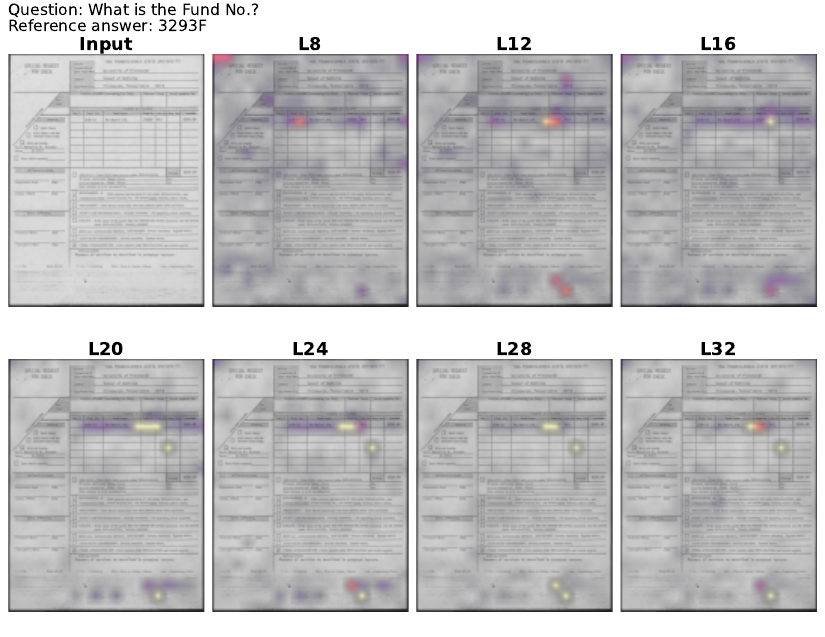}\par\nointerlineskip
\includegraphics[width=\linewidth]{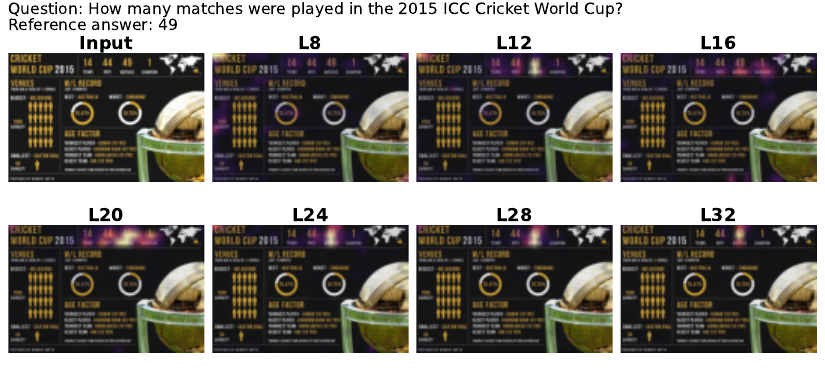}\par\nointerlineskip
\includegraphics[width=\linewidth]{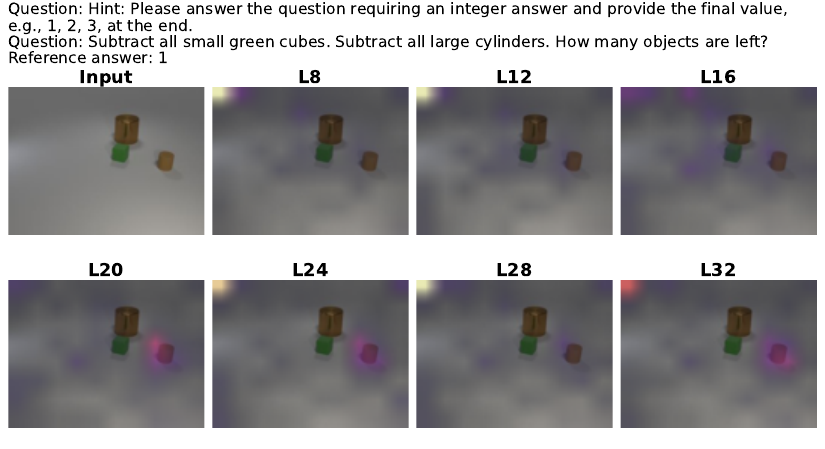}\par\nointerlineskip
\includegraphics[width=\linewidth]{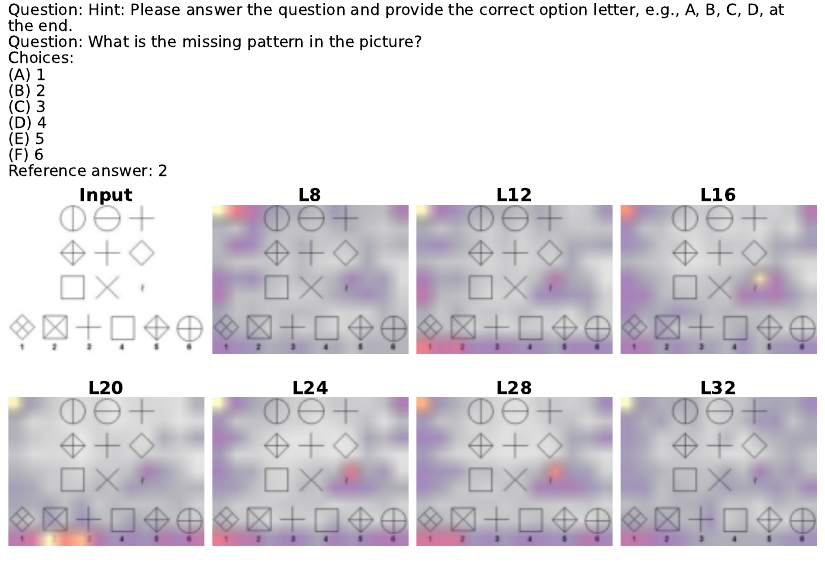}\par\nointerlineskip
\includegraphics[width=\linewidth]{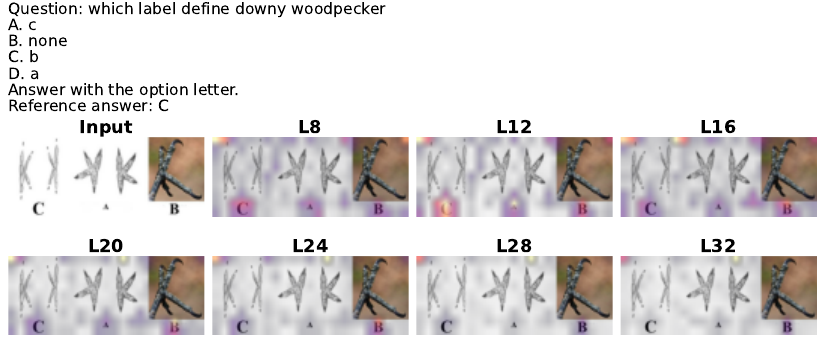}\par\nointerlineskip
\includegraphics[width=\linewidth]{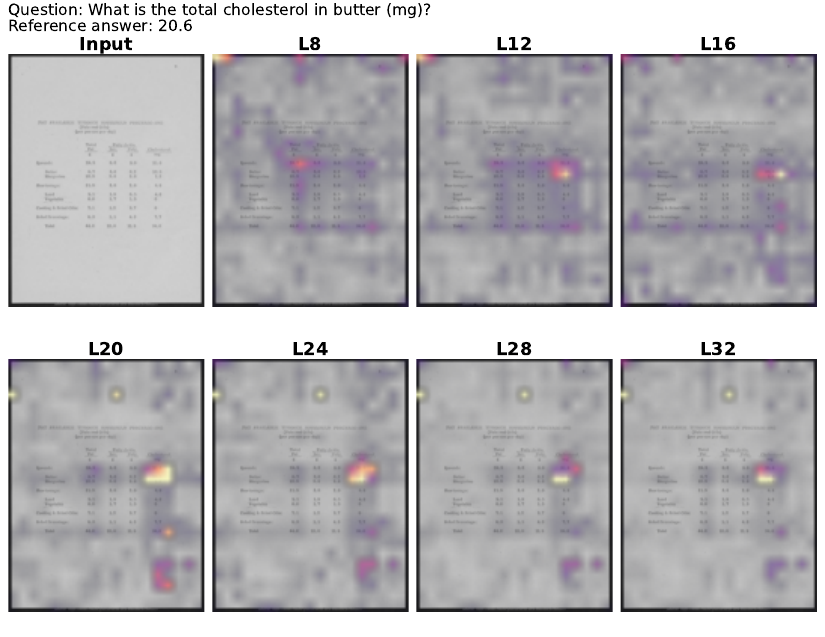}\par\nointerlineskip
\includegraphics[width=\linewidth]{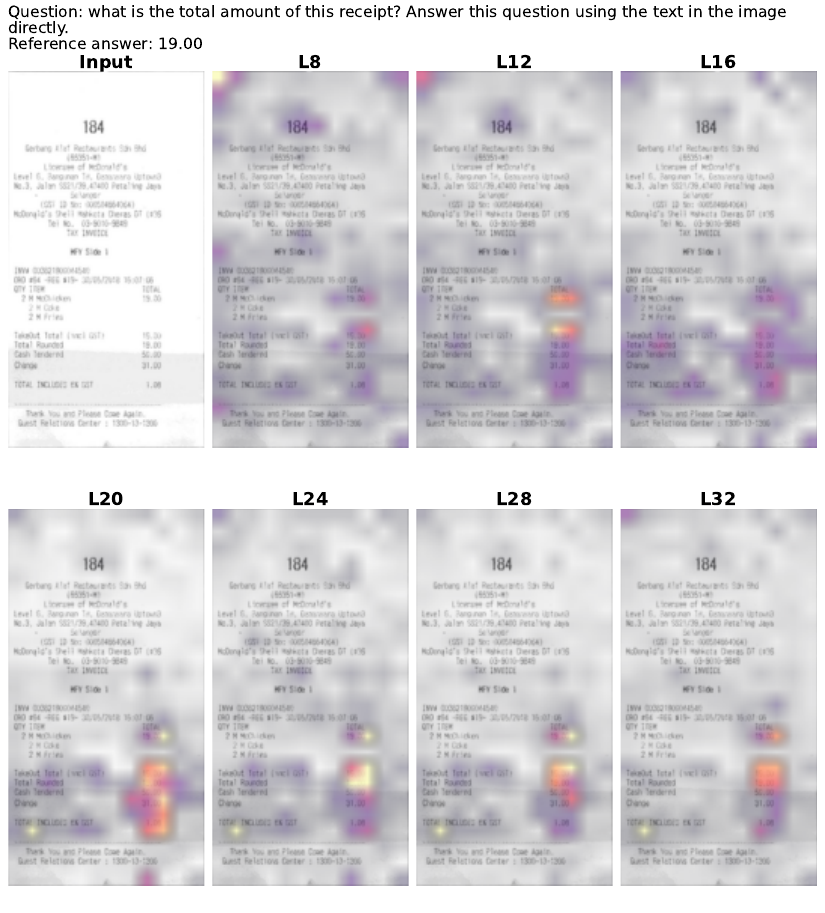}\par\nointerlineskip
\begin{minipage}{\linewidth}
\centering
\includegraphics[width=\linewidth]{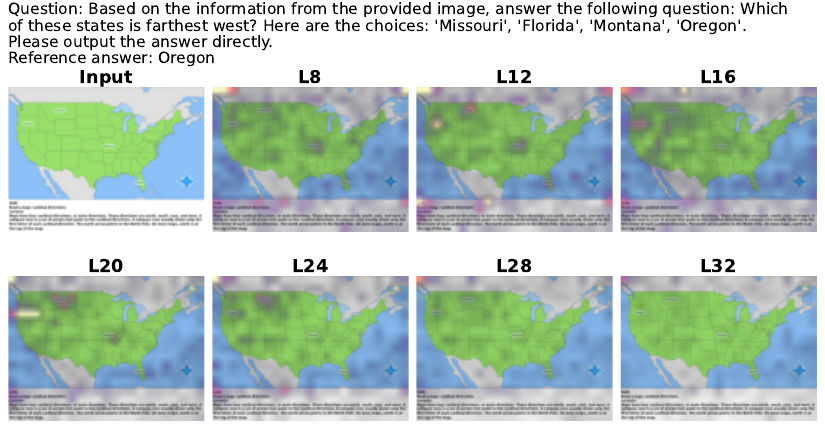}
\captionof{figure}{Layer-wise visual attention across nine examples, ordered as BLINK, DocVQA, InfoVQA, two MathVista~\citep{lu2023mathvista} examples, MMStar~\citep{MMStar}, two OCRBench examples, and OCRBench v2. Each example uses one shared scale across its displayed layers.}
\label{fig:layerwise-examples}
\end{minipage}
\endgroup

\end{document}